\documentclass[10pt,twocolumn,letterpaper]{article}

\usepackage[pagenumbers]{cvpr} 

\definecolor{cvprblue}{rgb}{0.21,0.49,0.74}
\usepackage[pagebackref,breaklinks,colorlinks,allcolors=cvprblue]{hyperref}

\usepackage{amssymb}
\usepackage{caption}
\usepackage{float}
\usepackage{multirow}

\def\paperID{4632} 
\def\confName{CVPR}
\def\confYear{2025}

\title{3DV-TON: Textured 3D-Guided Consistent Video Try-on via Diffusion Models}

\author{Min Wei$^{1,2}$~~~Chaohui Yu$^{\dagger1,2}$~~~Jingkai Zhou$^{1,2,3}$~~~Fan Wang$^{1}$\\ \small {$^1$DAMO Academy, Alibaba Group ~~~$^2$Hupan Lab~~~ $^3$Zhejiang University, }\\ \small \{weimin.wei, huakun.ych\}@alibaba-inc.com}

\begin{document}

\twocolumn[{%
    \maketitle
    \begin{figure}[H]
    \hsize=\textwidth 
    \centering
    \includegraphics[width=\textwidth]{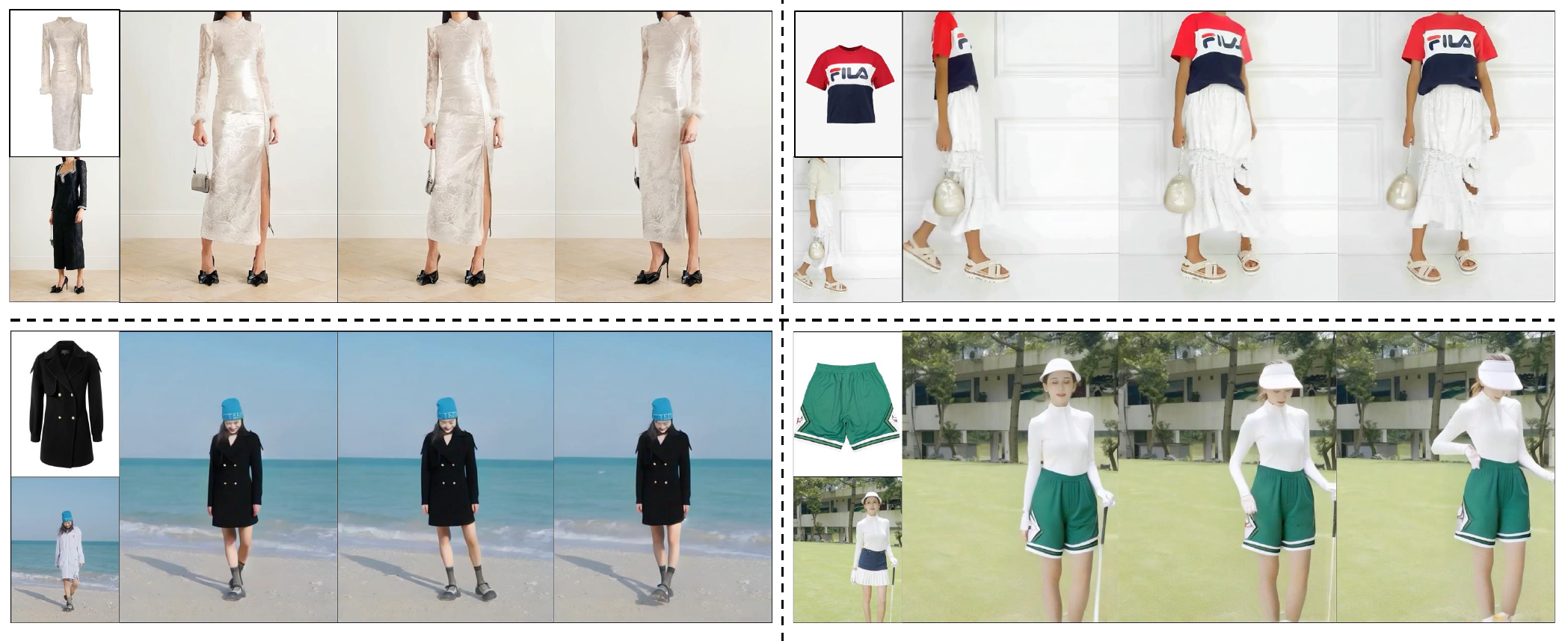}
    \caption{\textbf{Try-on videos generated by 3DV-TON.} Our method can handle various types of clothing and body poses, while accurately restoring clothing details and maintaining consistent texture motion.}
    \end{figure}
    \vspace{-0.15in}
}]

\maketitle

\renewcommand{\thefootnote}{}
\footnote{$^\dagger$ Corresponding author.}

\begin{abstract}
Video try-on replaces clothing in videos with target garments. Existing methods struggle to generate high-quality and temporally consistent results when handling complex clothing patterns and diverse body poses.
We present 3DV-TON, a novel diffusion-based framework for generating high-fidelity and temporally consistent video try-on results. Our approach employs generated animatable textured 3D meshes as explicit frame-level guidance, alleviating the issue of models over-focusing on appearance fidelity at the expanse of motion coherence.
This is achieved by enabling direct reference to consistent garment texture movements throughout video sequences.
The proposed method features an adaptive pipeline for generating dynamic 3D guidance: (1) selecting a keyframe for initial 2D image try-on, followed by (2) reconstructing and animating a textured 3D mesh synchronized with original video poses. We further introduce a robust rectangular masking strategy that successfully mitigates artifact propagation caused by leaking clothing information during dynamic human and garment movements.
To advance video try-on research, we introduce HR-VVT, a high-resolution benchmark dataset containing 130 videos with diverse clothing types and scenarios. Quantitative and qualitative results demonstrate our superior performance over existing methods. The project page is at this link~\url{https://2y7c3.github.io/3DV-TON/}
\end{abstract}

\section{Introduction}
\label{sec:intro}

Video try-on aims to change the person's clothing in the given video to a target garment, enabling customers to visualize themselves wearing clothing items without physical trials through enhanced immersion and interactivity. The process must preserve intricate garment details while maintaining consistent texture representation throughout the video sequence.

Prior video try-on works~\cite{fwgan,mvton,clothformer} typically employ flow-driven warping modules~\cite{vitonhd,dong2020fashion,ge2021parser,wang2018toward,xie2023gp} for precise garment alignment on human figures, complemented by neural generators to synthesize the final appearance.
However, these methods face inherent limitations from their reliance on warping operations: while effectively adapting garment geometry through shape deformation to match pose variations, they inherently compromise temporal coherence in generated sequences. This fundamental constraint hinders handling of substantial clothing deformations and complex occlusions, limiting practical application to simplified scenarios.

Recent advancements~\cite{vivid,tunneltryon} harness pre-trained diffusion models~\cite{sd,ddpm} to address limitations of conventional warping modules. These works~\cite{vivid, tunneltryon}  implement a dual-UNet architecture: a primary denoising UNet~\cite{unet} alongside a parallel reference UNet that directly extracts garment features, eliminating explicit warping.
Hierarchical temporal attention layers~\cite{animatediff} are integrated within the denoising net to model motion dynamics and mitigate inter-frame inconsistencies.
Concurrently, Diffusion Transformer (DiT)-based frameworks~\cite{dit} demonstrate enhanced performance in video try-on through superior generative scalability, as evidenced by works like~\cite{catv2ton,vitondit}. Nevertheless, empirical analysis in~\cite{videojam} reveals that pixel-reconstruction objectives in video diffusion models remain constrained in achieving robust temporal coherence.

\begin{figure}
    \centering
    \includegraphics[width=0.95\linewidth]{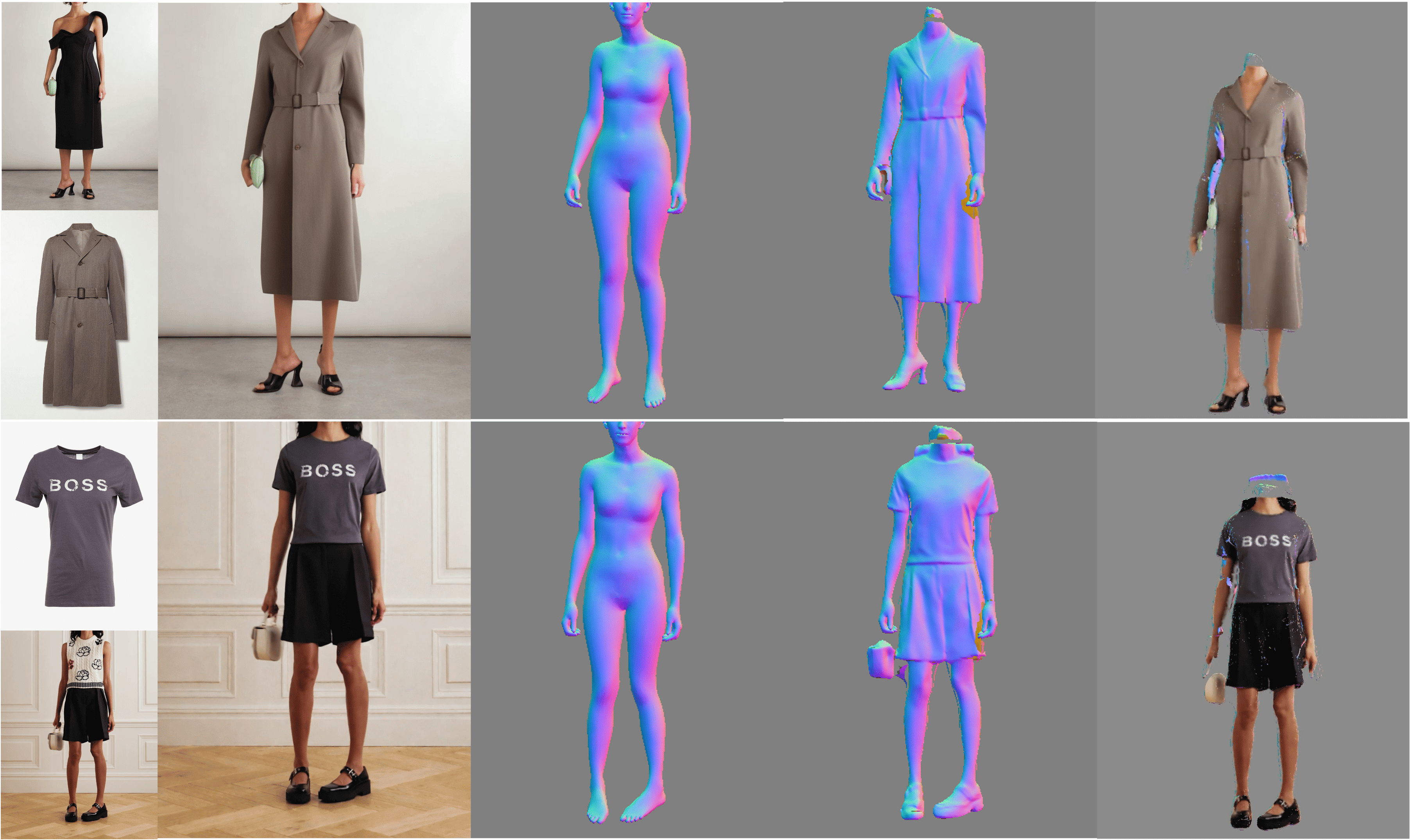}
    \caption{\textbf{Textured 3D guidance.} We construct the textured 3D guidance based on image try-on results, then animate the mesh after pasting the texture, providing a consistent texture motion reference on the appearance level.}
    \label{fig:3d_pipe}
    \vspace{-0.15in}
\end{figure}

In this paper, we present 3DV-TON, a diffusion-based framework for generating high-fidelity temporally-consistent video try-ons.
To tackle the limitation, models prioritize appearance fidelity over motion coherence, in prior literature where pixel-based reconstruction objectives inherently, we introduce explicit frame-level textured 3D guidance. 
Our method directly models 3D human meshes wearing target garments, ensuring spatiotemporal consistency across diverse poses and viewpoints through motion-aligned mesh propagation, which providing a consistent motion reference on the appearance level. While existing methods~\cite{champ,realisdance} employ 3D human priors, they exclusively utilize geometric structural cues without textured guidance. Our experiments demonstrate that geometric-only guidance (e.g., SMPL~\cite{smpl,smplx}) often fails to sufficiently constrain models, resulting in appearance-biased optimization and motion artifacts. Crucially, our textured 3D guidance uniquely preserves garment identity throughout video sequences, addressing a critical oversight in current video try-on works.

As illustrated in Figure~\ref{fig:3d_pipe}, our pipeline begins with selecting a frame through pose estimation, processed using advanced diffusion-based image try-on methods~\cite{catvton,ootd,idmvton}. This initial frame undergoes animatable textured 3D mesh reconstruction aligned with the source video motion to generate temporally consistent reference sequences. Unlike the previous warp module, our framework leverages single-image 3D reconstruction~\cite{icon,econ,sifu,lhm} to inherently establish spatiotemporal consistency, delivering robust appearance priors for the denoising UNet while reducing temporal attention dependencies. This strategy effectively bypasses complex warping operations through mesh animation, while benefiting from mature single-image reconstruction methods without task-specific retraining.

We further propose a dynamic rectangular masking strategy to prevent garment information leakage during human motion, which is a primary failure source in video try-on. To counter excessive masking, we implements both clothing images and try-on images as references to provide garments and environment context, and design an effective guidance feature extraction and fusion diffusion-based architecture. Comprehensive experiments demonstrate that our 3D-aware framework achieves superior visual quality and consistency in complex dynamic scenarios compared to existing approaches.

In summary, our main contributions are as follows:
\begin{itemize}[left=1em]
    \item We propose 3DV-TON, a novel diffusion-based video try-on method that employs textured 3D guidance to alleviate motion incoherence stemming from appearance bias. Our method effectively generates try-on videos maintaining consistent texture motion across varying body poses and camera viewpoints.
    \item We introduce a 3D guidance pipeline capable of adaptively generating animatable textured 3D meshes, ensuring consistent texture guidance across both spatial and temporal domains. The framework seamlessly integrates with existing methodologies without necessitating additional training.
    \item We establish a high-resolution video try-on benchmark enabling better evaluation of recent works, and demonstrate that our 3DV-TON outperforms existing video try-on methods in both quantitative and qualitative experiments.
\end{itemize}
\section{Related Works}
\label{sec:related}

\noindent \textbf{Image Virtual Try-on.}
Image virtual try-on aims to generate images of a target person wearing a given clothing. Many GAN-based methods~\cite{vitonhd,xie2023gp,ge2021parser,wang2018toward,dong2020fashion,huang2022towards,zhang2021pise,he2022style} typically first warp the clothing image onto the target person's body. Then, a generator is used to blend the warped clothing with the human body to produce realistic results. These methods rely on the accuracy of the warping module. Due to undesired distortions and artifacts caused by the TPS~\cite{tps}-based methods~\cite{minar2020cloth,viton}, many subsequent methods~\cite{xie2023gp,ge2021parser,huang2022towards,han2019clothflow,he2022style} have focused on predicting dense flow to achieve better warping of clothing, and have made significant progress. However, explicit warping techniques still struggle with complex poses and occlusions.

Recently, several works~\cite{ladivton,dcivton,ootd,idmvton,stablevton,catvton} tend to employ powerful pre-trained diffusion models~\cite{sd,ddpm} as an alternative to GANs~\cite{gan} to generate more realistic try-on results. 
OOTDiffusion and IDM-VTON~\cite{ootd,idmvton} utilized a dual-UNet~\cite{unet} structure and integrated clothing features and person features through self-attention. CatVTON~\cite{catvton} proposed to merge dual-UNet architectures, simplifying the training parameters and the inference process. However, applying image-based try-on techniques frame by frame to videos can lead to temporal inconsistent results.

\noindent \textbf{Video Virtual Try-on.} Compared to image try-on, video try-on needs to maintain temporal consistency between frames to generate realistic, high-quality results, which adds more challenges to the task. Previous works~\cite{fwgan,mvton,clothformer} typically employs a flow-based warping module~\cite{vitonhd,dong2020fashion,ge2021parser,wang2018toward,xie2023gp} for precise garment alignment on human bodys, and combine the warped clothing with the person in the video. In video try-on, warp-based methods also face challenges in handling complex textures and motion.

Recent diffusion-based works~\cite{vivid,tunneltryon,he2024wildvidfit,wang2024gpd}, build on the dual-UNet architecture, a primary denoising UNet~\cite{unet} alongside a parallel reference UNet that directly extracts garment features to preserve the visual quality, and insert hierarchical temporal modules~\cite{animatediff} to ensure temporal smoothness. 
ViViD~\cite{vivid} releaseed a new dataset and improves the generation resolution from 256 to 512. Some works~\cite{tunneltryon,wang2024gpd} utilized private datasets with a resolution of 512 and introduced techniques to emphasize the clothing. More recently, some works~\cite{vitondit,catv2ton} utilized the powerful diffusion transformer (DiT) framework and have made significant progress in the video try-on task.
However, these methods struggle to maintain consistent temporal coherence between frames, tend to generate over-smoothed deformed clothing textures. 

\noindent \textbf{Clothed 3D Human Reconstrction.} Previous works~\cite{diffavatar,de2023drapenet,gao2023cloth2tex,zhao2021m3d} typically requires modeling a 3D human body model and a clothing model, and then fit the clothing onto the human body model. Additionally, when animate the model, physical simulation is introduced to generate natural clothing movement. One line of research, \textit{e.g.}, DiffAvatar~\cite{diffavatar}, proposed a methods for body shape and garment assets recovery from 3D scan of a clothed person, and utilized differentiable simulation for co-optimizing garment and human body. Such methods have very high requirements for the input data. On the other hand, several methods reconstructs a clothed human from a single image and simultaneously models the clothing along with the person for animation. ICON~\cite{icon} used body-based normal estimation for implicit 3D reconstruction. ECON~\cite{econ} significantly improved reconstruction robustness by integrating explicit shape-based approaches with normal priors. SIFU~\cite{sifu} proposed a side-view conditioned implicit function to achieve more accurate reconstruction results. More recently, some works~\cite{weng2024template,lhm} introduced large reconstruction models (LRMs) to enable feed-forward clothed human reconstructions. These works are capable of generating photorealistic, animatable human avatars in seconds, but they struggle to produce flexible clothing motions.
\section{Method}

\begin{figure*}
    \centering
    \includegraphics[width=0.95\linewidth]{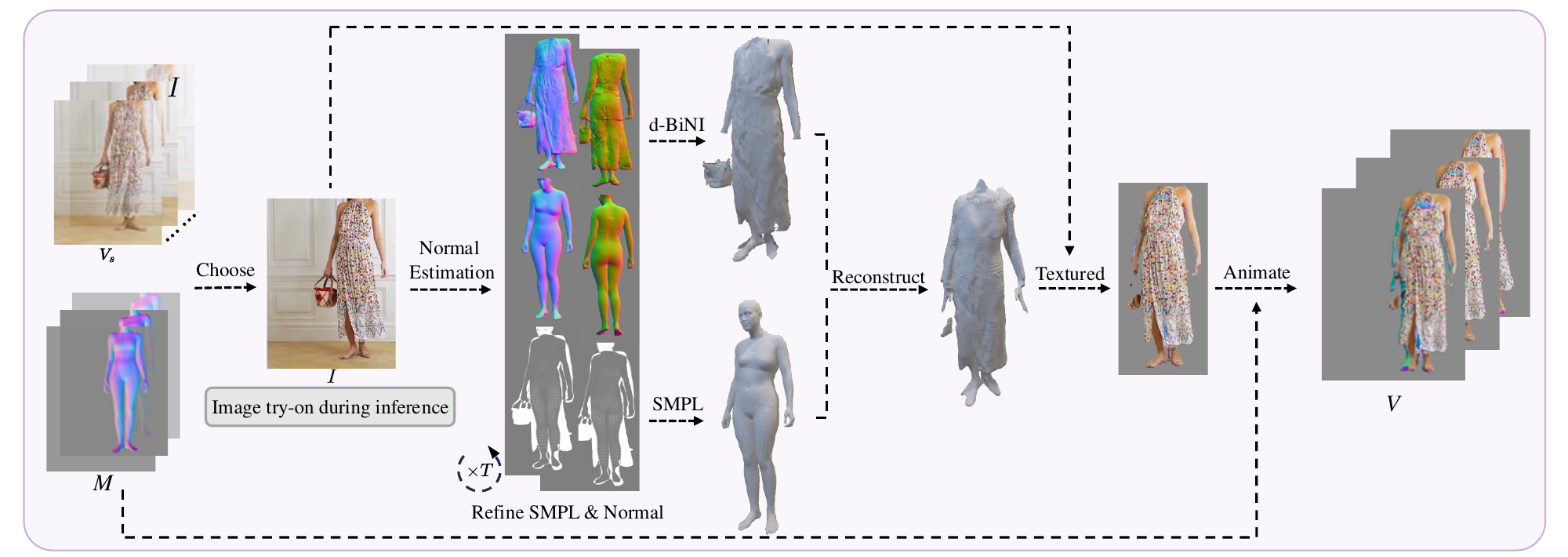}
    \label{fig:arch_3d}
\end{figure*}

\begin{figure*}
    \centering
    \includegraphics[width=0.95\linewidth]{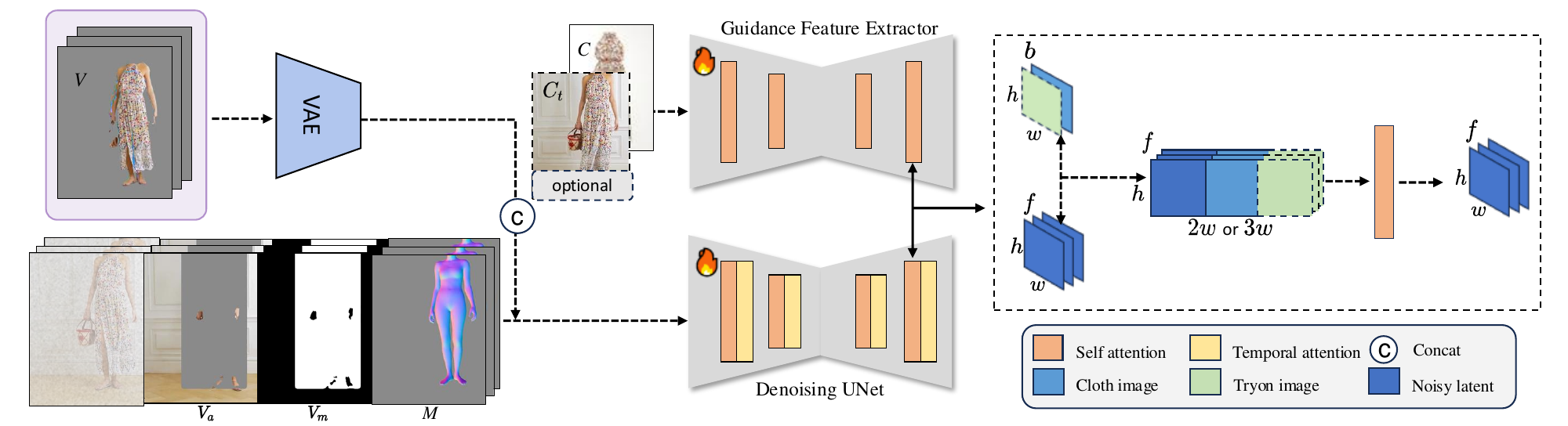}
    \caption{\textbf{The overview of 3DV-TON.} Given a video, we first use our 3D guidance pipeline to select a frame $I$ adaptively, then reconstruct a textured 3D guidance and animate it align with the original video, \textit{i.e.} $V$. We employ a guidance feature extractor for the clothing image $C$ and the try-on images $C_t$, and perform feature fusion using the self-attentions in the denoising UNet.}
    \label{fig:arch}
\end{figure*}

The overview pipeline of our 3DV-TON is illustrated in Figure~\ref{fig:arch}. We first introduce the textured 3D guidance generation pipeline in Section~\ref{sec:3dguidance}. Then, the model architecture and training strategy are illustrated in Section~\ref{sec:netarchi}.

\subsection{Animatable Textured 3D Guidance}
\label{sec:3dguidance}
\noindent \textbf{SMPL\&SMPLX.} The Skinned Multi-Person Linear (SMPL) model~\cite{smpl} is a 3D parametric human model that defines the shape topology of body. It uses shape parameters $\beta \in \mathbb{R}^{10}$ and pose parameters $\theta \in \mathbb{R}^{24\times 3}$ to represent the 3D human body mesh $M(\beta, \theta)$ as:
\begin{align}
    T_p(\beta, \theta) &= \overline{T} + B_s(\beta) + B_p(\theta) \nonumber,\\
    M(\beta, \theta) &= W(T_p(\beta, \theta), J(\beta), \theta, \mathcal{W}),
\end{align}
where $\overline{T}$ is the mean template shape, $B_s(\beta), B_p(\theta)$ are vectors of vertices representing offsets from the template. $T_p(\beta, \theta)$ is the non-rigid deformation from $\overline{T}$. $W(\cdot)$ is the linear blend skinning (LBS)~\cite{lbs} function applied to rotate the vertices around the joint center $J(\beta)$ with the smoothing defined by the blend weights $\mathcal{W}$.

The SMPL-X model~\cite{smplx} builds upon SMPL, adding features for hands and face, enhanceing facial expressions, finger movements, and detailed body poses.

\noindent \textbf{Clothed Human Reconstruction\&Animation.} Our human reconstruction method is based on ECON~\cite{econ}. Given a video, 
we choose a frame $I$ according to estimated body pose adaptively, which performing image try-on~\cite{catvton,idmvton,ootd} during inference, as the input of normal estimation network~\cite{sapiens, icon, econ}.
To guide the normal map prediction for clothed normal map (denoted as $\hat{\mathcal{N}}^c_{\{F,B\}}$ where $F, B$ denote front/back view), and ensure robustness across poses, we use body normal maps $\mathcal{N}^b_{\{F,B\}}$ rendered from the estimated SMPL-X $M^b(\beta, \theta)$ as reconstruction conditions. Accurate alignment between body estimation and clothing silhouettes proves crucial for this process. However, existing human pose and shape (HPS) regressors~\cite{hybrikx, hybrik, gvhmr,wham} fails to provide pixel-aligned SMPL-X fits. 
Unlike previous works~\cite{econ,icon} that require precise body pose optimization, our method prioritizes clothing reconstruction accuracy over anatomical details. By eliminating SMPL-X pose $\theta$ optimization during parameter refinement, we reduce optimization steps and reconstruction time to $\sim$30s while maintaining performance. We additionally optimize camera scale $s$ to address systematic camera estimation errors in HPS methods.
Our optimization process initializes with estimated SMPL-X's shape $\beta$, translation $t$ parameters and camera scale $s$, focusing on minimizing silhouette and normal loss:
\begin{align}
    &\mathcal{L}_{\text{SMPL-X}} = \mathcal{L}_{N_d} + \mathcal{L}_{S_d} + \lambda \cdot \min(\mathbf{d} - s,0),\\
    &\mathcal{L}_{N_d} = \lvert\hat{\mathcal{N}}^c - \mathcal{N}^b(\beta,t,s)\rvert, \quad \mathcal{L}_{S_d} = \lvert\hat{\mathcal{S}}^c - \mathcal{S}^b(\beta,t,s)\rvert \nonumber,
\end{align}
where $\mathcal{L}_{N_d}$ is a normal map L1 loss, $\mathcal{L}_{S_d}$ is a L1 loss between the silhouettes of the SMPL-X $\mathcal{S}^b$ and the clothed human mask $\hat{\mathcal{S}^c}$ segmented from image $I$.
We additionally introduce a unidirectional regularization penalty to address frequent partial body observations in training data (see \textit{Supplementary Materials}.), activated during loss computation when camera scale falls below the dataset-defined threshold $\mathbf{d}$. Following SMPL-X refinement, we iteratively updating the normal map and SMPL-X parameters through $T$ refinement cycles.

We reconstruct the front and back surface using depth-aware silhouette-consistent bilateral normal integration (d-BiNI) method introduced by \cite{econ, bini}. However, poses often result in self-occlusions, which cause large portions of the surfaces to be missing. In such cases, we use a simple way to infill the missing surface using the estimated SMPL-X body that invisible to front or back cameras, and union the parts of surface by surface reconstruction methods~\cite{psr,nksr}. Since the reconstructed mesh is aligned with the image pixels, we can simply use interpolated pixel values as the mesh texture after calculating visibility. For the invisible body areas, we use normals as the texture.

The reconstructed clothed human inherit the hierarchical skeleton and skinning weights from the underling SMPL-X body model, allowing to animate it using the estimated SMPL poses~\cite{gvhmr} from the original video. Specifically, for each vertices $j$ of the clothed human mesh, we use k-nearest neighbor (KNN) search to obtain a set $\mathcal{K}_j$ composed of $K$ neighboring control points denoted as $\{p_k|k\in\mathcal{K}_j\}$ in canonical SMPL-X model. Then, the interpolation weights for control points $p_k$ can be computed as:
\begin{align}
    &w_{jk} = \frac{\hat{w}_{jk}}{\sum_{k\in\mathcal{K}_j}\hat{w}_{jk}}, \quad \hat{w}_{jk} = \exp(-d^2_{jk}),
\end{align}
where $d_{jk}$ is the distance between vertices $j$ and the neighboring vertices $p_k$ in SMPL-X.
The overall 3D guidance generation pipeline is depicted in the upper part of Figure~\ref{fig:arch}.

\subsection{Network Architecture}
\label{sec:netarchi}
\noindent \textbf{Controlled Diffusion Model.} Stable Diffusion~\cite{ddpm,sd} is the basis for our network that consists of a varialtional autoencoder (VAE)~\cite{vae} and a denoising UNet~\cite{unet}. 
Given an image $x_0$ and a control condition $\mathbf{c}$, the VAE first encodes the image $x_0$ into latent space: $z_0 = \mathcal{E}(x_0)$. The UNet learns to predict a noise $\epsilon_\theta$ or velocity $v_\theta$ based on the control condition $\mathbf{c}$ and the noisy latent $z_t: z_t = \alpha_t z_0 + \sigma_t \epsilon$. The training loss of the UNet can be formulated as:
\begin{align}
    \mathcal{L}_{LDM} &= \mathbb{E}_{z,c,\epsilon,t}[\lVert v_t - v_\theta(z_t,t,\mathbf{c})\rVert_2^2],
\end{align}
where $t$ represent the diffusion timestep, $\epsilon \sim \mathcal{U}(0,I)$, $v_t = \alpha_t \epsilon - \sigma_t z_0$~\cite{vpred}. In inference, data samples can be generated from Gaussian noise $z_T \sim \mathcal{N}(0, I)$ by the denoising process. 

\noindent \textbf{Guidance Feature Extractor.} Our method employs two reference conditions: clothing images $C\in\mathbb{R}^{b\times 3\times H\times W}$ and try-on images $C_t\in\mathbb{R}^{b\times 3\times H\times W}$ encoded into latent space through VAE encoder $\mathcal{E}$ as $\mathbf{C} = \mathcal{E}(C)$ and $\mathbf{C_t}=\mathcal{E}(C_t)$.
These latent representations ($\mathbf{C}, \mathbf{C_{t}}\in \mathbb{R}^{b\times 4\times h\times w}$ are concatenated along the batch dimension to form composite reference features $\mathbf{F}\in \mathbb{R}^{2b\times 4\times h\times w}$. We duplicate the denoising UNet as the Guidance Feature Extractor that capture the visual features of the clothing images and try-on images. Note that we remove text encoders and all cross attention layers cause our textured 3D guidance provided sufficiently explicit visual reference. 

\noindent \textbf{Denoising Network.} We employ a UNet architecture from Stable Diffusion~\cite{sd} without cross-attention layers, extended into a pseudo-3D structure through temporal module~\cite{animatediff} integration to enable realistic motion generation, serving as our base denoising network. 
Given a batch of source videos $V_s\in\mathbb{R}^{b\times3\times f\times H\times W}$, with corresponding clothing-agnostic videos $V_a \in \mathbb{R}^{b\times 3\times f\times H\times W}$ and mask videos $V_m \in \mathbb{R}^{b\times 1\times f\times H\times W}$. We extimate the SMPL sequences $M$ using HPS methods~\cite{wham,gvhmr}. Our adaptive 3D guidance pipeline then generates textured 3D guidance $V$. The denoising input comprises concatenated features along the channel dimension: the noisy latent video $z_t$, the latent clothing-agnostic video $\mathcal{E}(V_a)$, the resized mask video $V'_m$, the SMPL geometic guidance $\mathcal{E}(M)$ and the textured 3D guidance $\mathcal{E}(v)$. To accommodate this $17$-channel input, we expand the UNet's initial convolutional layer with zero-initialized weights.

Our guidance feature extractor avoids feature fusion between clothing and try-on images. Instead, we implement texture-aware fusion through spatial attention mechanisms (Figure~\ref{fig:arch}).
For each latent $x_s^i \in \mathbb{R}^{b\times c\times f\times h\times w}$ entering the $i$-th self-attention layer, we retrieve corresponding reference features corresponding reference feature $x_f^i \in \mathbb{R}^{2b\times c\times h\times w}$ from the extractor. These features split into clothing feature $x_c^i$ and the try-on feature $x_{c_t}^i$, which we temporally align by replicating along the frame dimension to obtain $\hat{x}_c^i,\hat{x}_{c_t}^i \in \mathbb{R}^{b\times c\times f\times h\times w}$. As shown in Figure~\ref{fig:arch}, the three types of features are concatenated along the spatial dimensions, denoted by $\hat{x}_s^i \in \mathbb{R}^{b\times c\times f\times h\times 3w}$, . 
Then feature fusion is performed by the attention layer of the denoising network to obtain the latent $x_s^{i+1}$, which incorporates both the fine clothing textures and the frame-consistent 3D features.

\noindent \textbf{Training Strategy.} Inspired by \cite{moviegen, vivid, magicanimate}, our model is trained on both image and video datasets by treating images as single-frame videos. 
During training, we randomly select a type of dataset via a random number $r\sim \mathcal{U}(0,1)$, where $\mathcal{U}(\cdot, \cdot)$ is the uniform distribution. If $r < \tau$, we use the sampled images for training, and set the gradients of the temporal attention as zero to freeze the temporal module. Otherwise, we sample data from the video dataset, and make the temporal attention trainable. Hence, our training objective can be formulated as:
\begin{align}
    \mathcal{L} = \mathbb{E}_{z,\epsilon,t}[\lVert v_t - v_\theta(z_t,t,\mathbf{C},\mathbf{C}_t,\mathbf{V}) \rVert], ~ z\sim \{z_{img}, z_{vid}^{1:f}\}.
\end{align}
We incorporate control conditions through Classifier-Free Guidance (CFG)~\cite{cfg}. Specifically, we randomly omit the clothing image $\mathbf{C}$ with a probability of $p_1$, the try-on image $\mathbf{C_t}$ with a probability of $p_2$, and the textured 3D guidance $\mathbf{V}$ with a probability of $p_3$.

\noindent \textbf{Masking Strategy.} Our pipeline begins with garment segmentation using either human parsing~\cite{li2020self} or segmentation model~\cite{kirillov2023segment} to generate clothing masks. We compute bounding boxes from these masks and employ a human estimation model~\cite{hybrik,hybrikx,guler2018densepose}to selectively critical anatomical regions (\textit{e.g.} face and hands) while preserving body detail. This streamlined approach effectively prevents garment transfer failures caused by leaking clothing. Please refer to Section~\ref{sec:ab} for illustration.

\begin{figure}[tbp]
    \centering
    \tabcolsep=0.02in
    \begin{tabular}{ccccc}

        \multicolumn{5}{c}{
            \includegraphics[width=\linewidth]{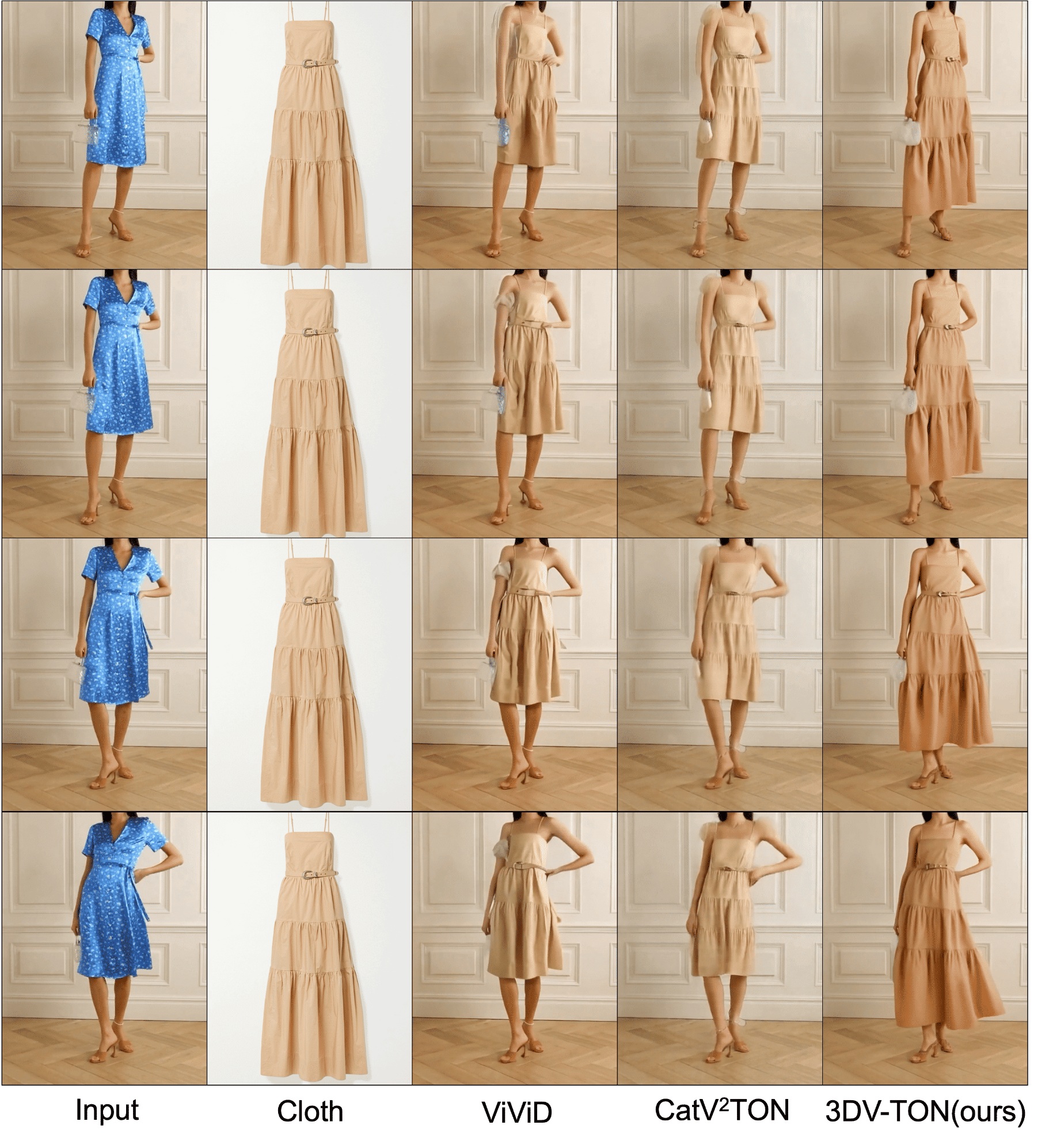}
        }\\
    \end{tabular}
    \caption{\textbf{Qualitative comparison for dress try-on on the ViViD dataset.}}
    \label{fig:qu_vivid_dress}
\end{figure}

\section{Experiments}
\subsection{Datasets}

\noindent \textbf{Training datasets.} We use two image datasets, VITON-HD\cite{vitonhd} and DressCode~\cite{dresscode}, along with one video dataset, ViViD~\cite{vivid}, to train our diffusion model. 
Due to the low resolution of the VVT~\cite{fwgan} dataset, we opt not to use it for training. 
Specifically, the VITON-HD dataset contains 13,678 images of upper-body clothing with corresponding model images. The DressCode dataset includes 15,363 images of upper-body clothing, 8,951 images of lower-body clothing, and 2,947 images of dresses, along with images of models wearing these garments. The ViViD dataset consists of 9,700 videos featuring models along with corresponding clothing images. This dataset contains 4,823 video-image pairs for upper-body clothing, 2,133 for lower-body clothing, and 2,744 for dresses, with a total of 1,213,694 frames. All image and videos are resized to $768\times576$ for training. For video data, we randomly select a frame to construct the try-on image condition using the image try-on method~\cite{catvton,ootd,idmvton}. For image data, we set all try-on conditions to be empty.

\noindent \textbf{HR-VVT benchmark.} Owing to the limitations of the ViViD dataset, which contains limited scenarios, and VVT dataset, which only includes upper-body clothing and exhibits relatively uniform body poses, coupled with a low resolution of only $256\times192$, it is challenging to accurately assess video try-on methods.
Therefore, we have constructed a high-resolution ($\sim$720p) video try-on benchmark called HR-VVT that includes 130 videos with 50 upper-body clothing, 40 lower-body clothing, and 40 dresses, with a variety of garments and motions in complex scenarios. Please refer to our \textit{Supplementary Materials} for more dataset details.

\subsection{Implementation Details}
\noindent \textbf{Textured 3D guidance.} During the 3D human reconstruction process, to achieve better body estimation, we use a image-based HPS regressor~\cite{hybrikx,pymaf, pixie} with SMPL-X and iteratively refined the SMPL-X parameters $T=10$ times for clothed human reconstruction based on ECON~\cite{econ}. To ensure smooth animation of the 3D guidance, we employ a video-based SMPL estimation approach~\cite{wham, gvhmr}. Due to the differences between image-based and video-based estimation methods, we use the body from the video-based estimation as the binding template to avoid texture distortion and animate it to render a guidance video, which is aligned with the source video. Please refer to \textit{Supplementary Materials} for more details.

\noindent \textbf{Training.} We initialize our guidance feature extractor and denoising network using the weights from SD1.5~\cite{sd}, and employ Animatediff~\cite{animatediff} to initialize the temporal attention. Our model is trained in a single-stage manner using $768\times576$ resolution, 32 frames (2 strides for videos) data. The model was trained using A800 GPUs for $40000$ steps with a learning rate of 1e-5. 

\begin{figure}[tbp]
    \centering
    \tabcolsep=0.02in
    \begin{tabular}{ccccc}

        \multicolumn{5}{c}{
            \includegraphics[width=\linewidth]{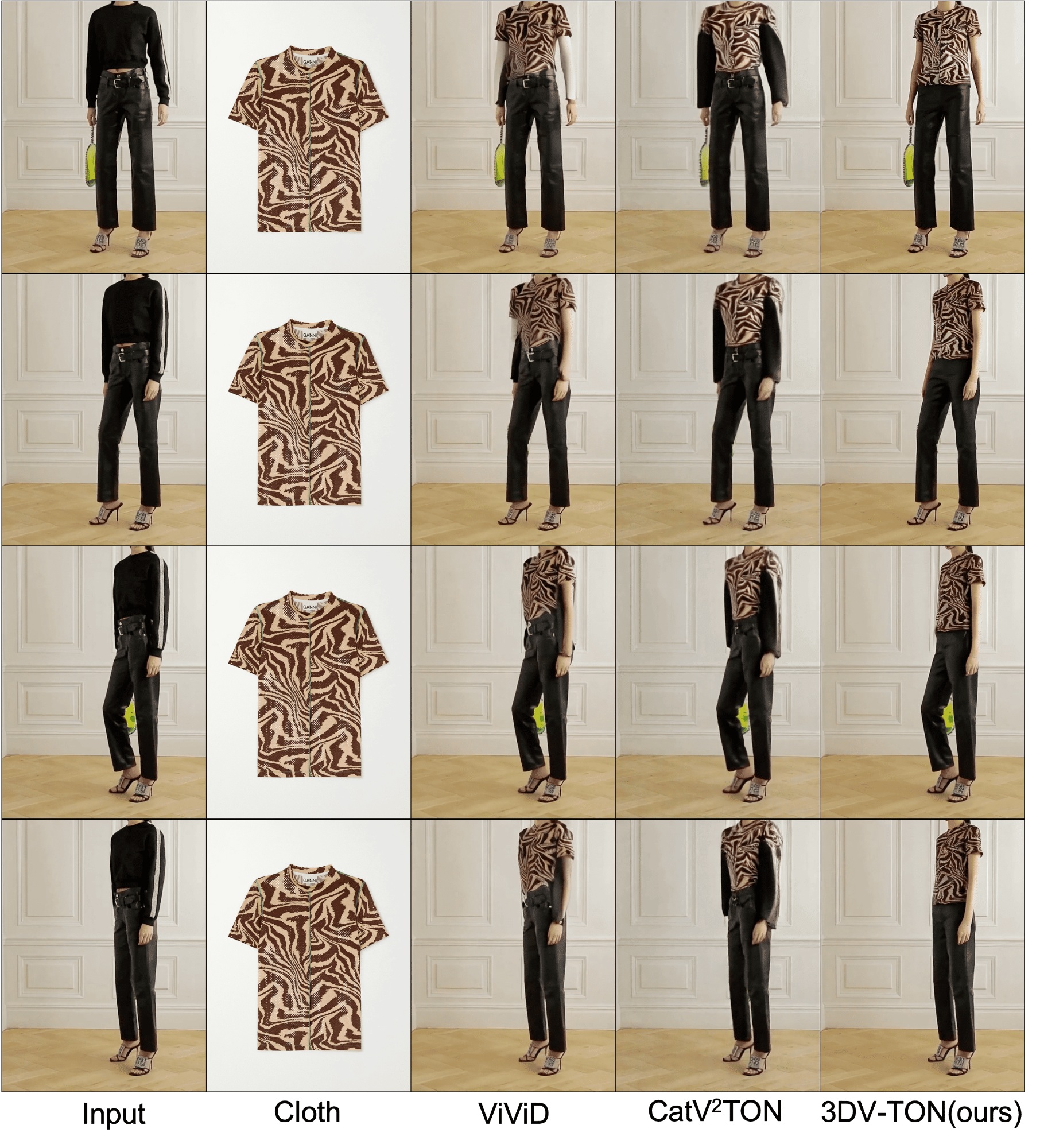}
        }\\
    \end{tabular}
    \caption{\textbf{Qualitative comparison for upper garment try-on on the ViViD dataset.}}
    \label{fig:qu_vivid_upper}
\end{figure}

\begin{figure}[tbp]
    \centering
    \tabcolsep=0.02in
    \begin{tabular}{ccccc}

        \multicolumn{5}{c}{
            \includegraphics[width=\linewidth]{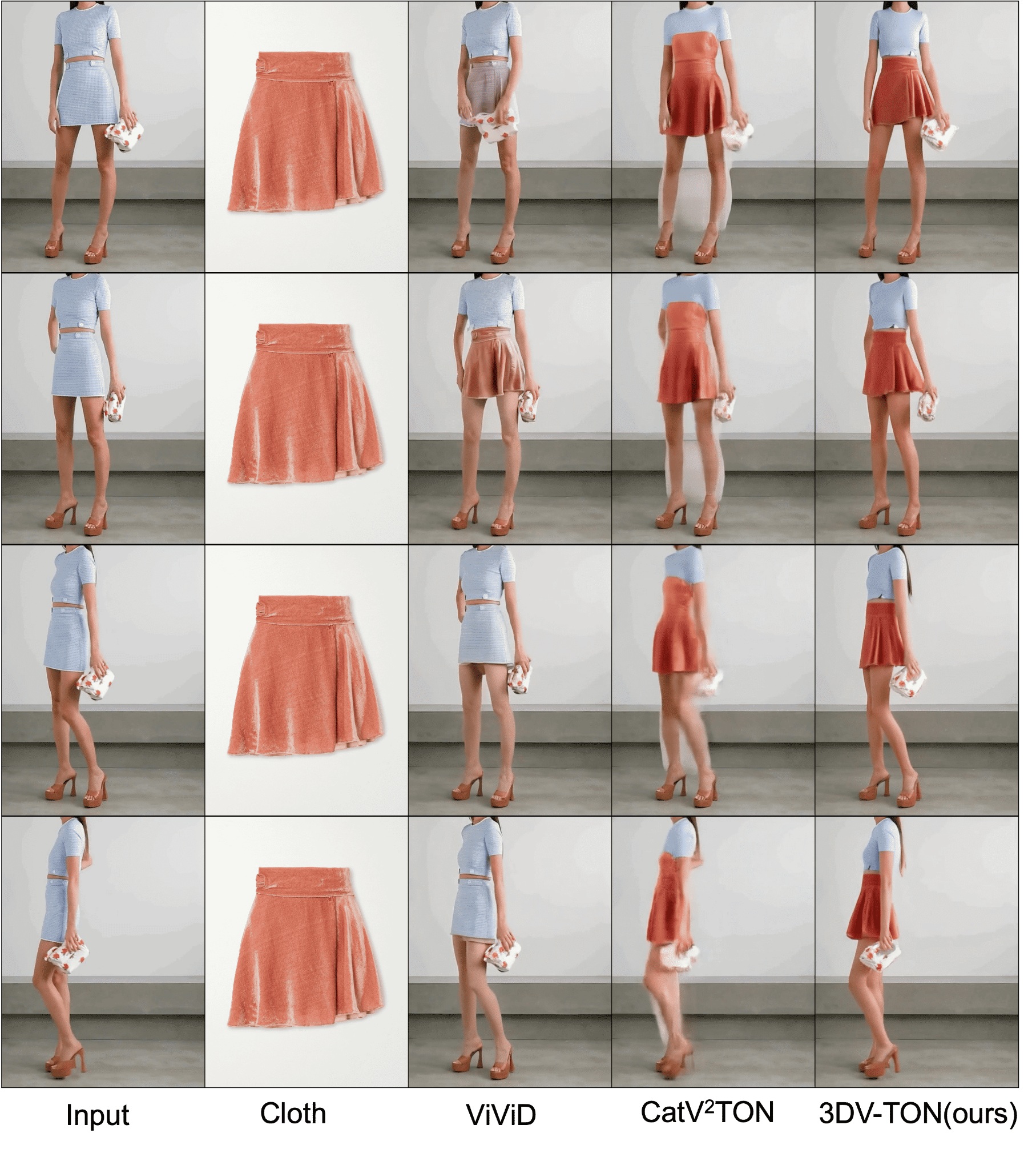}
        }\\
    \end{tabular}
    \caption{\textbf{Comparison for lower garment try-on on ViViD.}}
    \label{fig:qu_vivid_lower}
\end{figure}

\begin{figure}[tbp]
    \centering
    \tabcolsep=0.02in
    \begin{tabular}{ccccc}

        \multicolumn{5}{c}{
            \includegraphics[width=\linewidth]{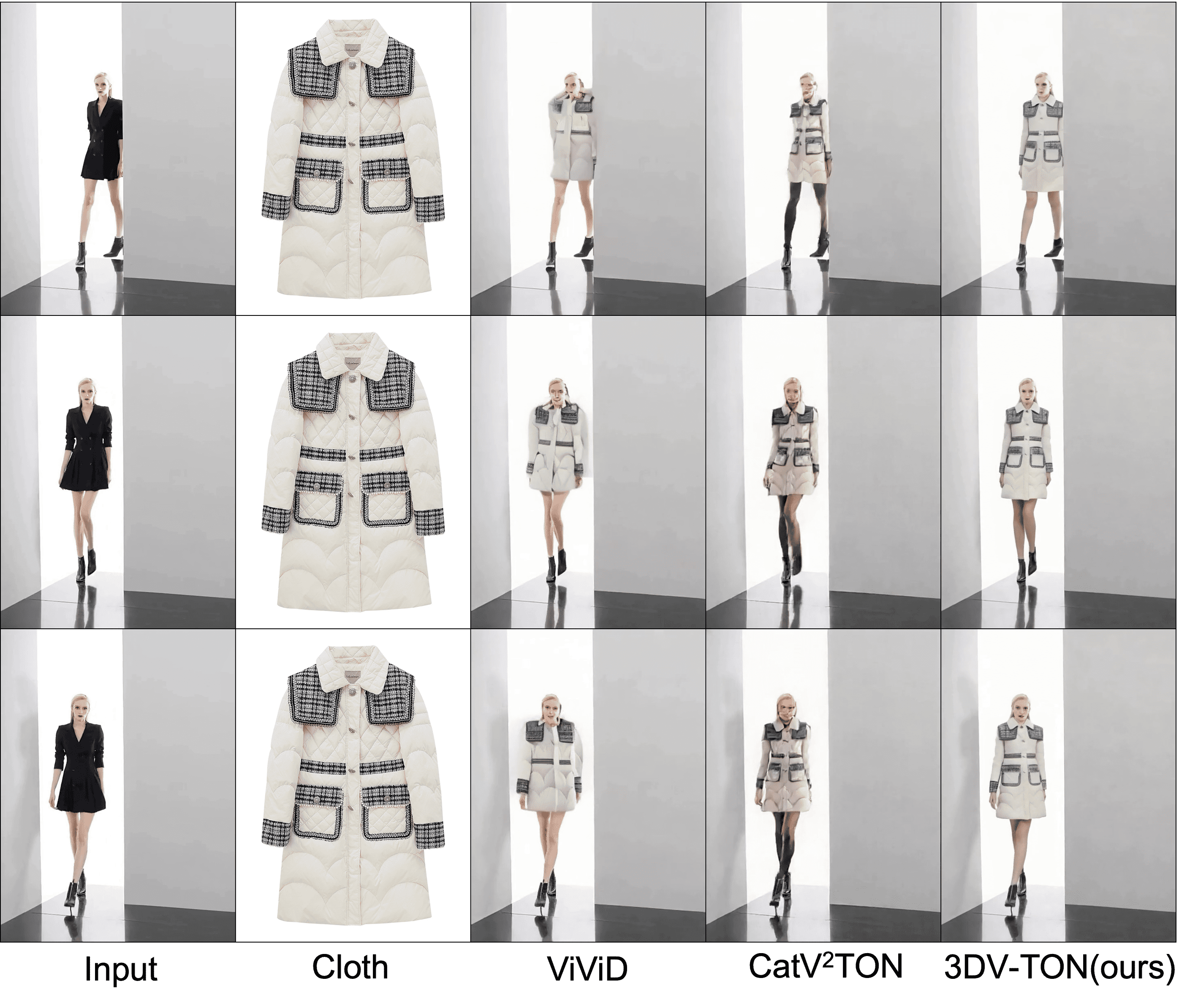}
        }
    \end{tabular}
    \caption{\textbf{Qualitative comparison for dress try-on on HR-VVT.}}
    \label{fig:qu_taobao_dress}
\end{figure}

\begin{figure}[tbp]
    \centering
    \tabcolsep=0.02in
    \begin{tabular}{ccccc}

        \multicolumn{5}{c}{
            \includegraphics[width=\linewidth]{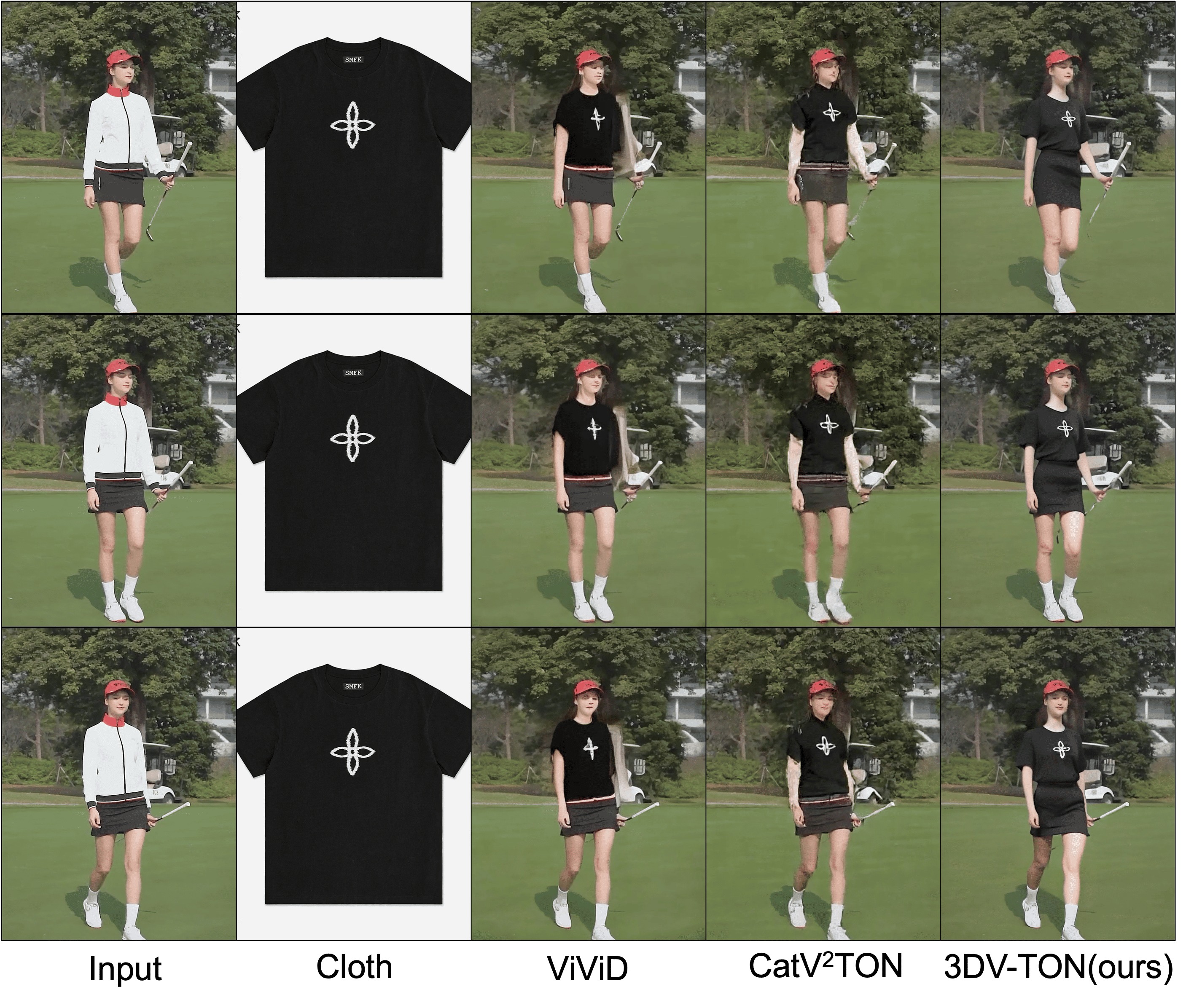}
        }
    \end{tabular}
    \caption{\textbf{Comparison on upper garment try-on on HR-VVT.}}
    \label{fig:qu_taobao_upper}
\end{figure}

\begin{figure}[tbp]
    \centering
    \tabcolsep=0.02in
    \begin{tabular}{ccccc}

        \multicolumn{5}{c}{
            \includegraphics[width=\linewidth]{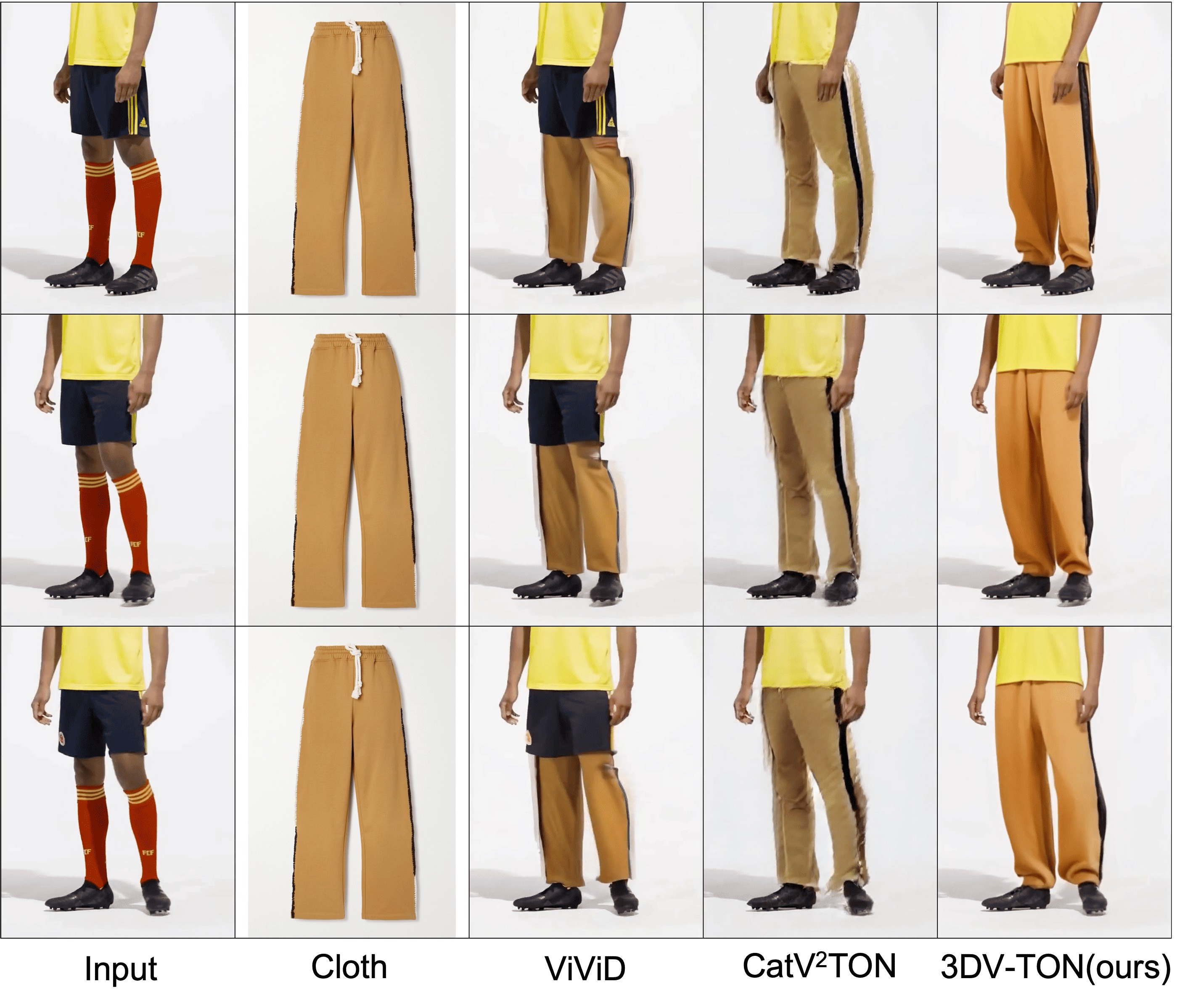}
        }
    \end{tabular}
    \caption{\textbf{Comparison on lower garment try-on on HR-VVT.}}
    \label{fig:qu_taobao_lower}
\end{figure}

\begin{table}[tbp]
\centering
    \resizebox{\linewidth}{!}{
    \begin{tabular}{l|cccc|cc}
    \hline
    \multirow{2}{*}{Method} & \multicolumn{4}{c}{Paired} & \multicolumn{2}{c}{Unpaired}\\
    &SSIM$\uparrow$ & LPIPS$\downarrow$ & $VFID_{I3D}$$\downarrow$ & $VFID_{RexNeXt}$$\downarrow$ & $VFID_{I3D}$$\downarrow$ & $VFID_{RexNeXt}$$\downarrow$\\
    \hline
    
    StableVITON~\cite{stablevton} & 0.8019 & 0.1338 & 34.2446 & 0.7735 & 36.8985 & 0.9064\\
    OOTDiffusion~\cite{ootd} & 0.8087 & 0.1232 & 29.5253 & 3.9372 & 35.3170 & 5.7078\\
    IDM-VTON~\cite{idmvton} & 0.8227 & 0.1163 & 20.0812 & 0.3674 & 25.4972 & 0.7167\\
    \hline
    StableVITON+AM~\cite{magicanimate} & 0.8207 & 0.1291 & 19.9239 & 0.7586 & 22.0262 & 0.8283\\
    OOTDiffusion+AM~\cite{magicanimate} & 0.8154 & 0.1244 & 19.3173 & 0.9382 & 23.3938 & 1.1485\\
    IDM-VTON+AM~\cite{magicanimate} & 0.8252 & 0.1212 & 18.2048 & 0.4481 & 22.5881 & 0.5397\\
    \hline
    ViViD~\cite{vivid} & 0.8029 & 0.1221 & 17.2924 & 0.6209 & 21.8032 & 0.8212\\
    CatV$^2$TON~\cite{catv2ton} & \underline{0.8727} & \underline{0.0639} & 13.5962 & 0.2963 & 19.5131 & 0.5283\\
    3DV-TON (Ours) & 0.8681 & 0.0707 & \underline{13.4062} & \underline{0.2741} & \underline{19.4714} & \underline{0.3664}\\
    \textit{3DV-TON}$^*$ (Ours) & \textbf{0.8992} & \textbf{0.0521} & \textbf{10.9680} & \textbf{0.2033} & \textbf{18.1151} & \textbf{0.3149} \\
    \hline

    \end{tabular}}

    \caption{\textbf{Quantitative comparison on the ViViD dataset.} $^*$ indicates our method using the same mask with ViViD~\cite{vivid}. }
    \label{tab:vvt}
\end{table}

\subsection{Qualitative Results} 
We conduct qualitative comparisons with the currently available method for video try-on that released the inference code, ViViD~\cite{vivid} and CatV$^2$TON\cite{catv2ton}. The clothing images and the person videos are from ViViD-S~\cite{vivid,catv2ton} test set and our HR-VVT set. None of the images or videos have appeared in the training data. 

\noindent \textbf{Comparisons on ViViD dataset.} As shown in Figure~\ref{fig:qu_vivid_dress} and Figure~\ref{fig:qu_vivid_upper}, other methods suffer from artifacts and generate garments limited by the patterns of the original clothing, while our method generates accurate garment shapes, offers better visual quality, and produces realistic clothing motion that adapts to the person's motions. Figure~\ref{fig:qu_vivid_lower} shows that ViViD~\cite{vivid} fails at this case, while CatV$^2$TON generates inconrrect gaments along with blurriness and artifacts. In contrast, Our 3DV-TON generates accurate clothing with good temporal consistency.

\noindent \textbf{Comparisons on HR-VVT benchmark.} Our HR-VVT benchmark includes a more diverse set of environments, clothing. As shown in Figure~\ref{fig:qu_taobao_dress}, ViViD~\cite{vivid} struggles with perspective changes during subject movement, and CatV$^2$TON fails to preserve garment consistency. In contrast, our method leverages explicit textured 3D guidance to maintain visual coherence across viewpoints and motion sequences. Figure~\ref{fig:qu_taobao_upper} shows that our approach's superiority in outdoor scenarios, where  competing methods exhibit artifacts and unrealistic texturing. Figure~\ref{fig:qu_taobao_lower} demonstrates how our robust 3D guidance pipeline ensures reliable performance even with partial character visibility. Please refer to our \href{https://2y7c3.github.io/3DV-TON/}{\textit{Project Page}} for more qualitative comparisons and video results.

\begin{table}[tbp]
\centering
    \resizebox{\linewidth}{!}{
    \begin{tabular}{l|cccc|cc}
    \hline
    \multirow{2}{*}{Method} & \multicolumn{4}{c}{Paired} & \multicolumn{2}{c}{Unpaired}\\
    &SSIM$\uparrow$ & LPIPS$\downarrow$ & $VFID_{I3D}$$\downarrow$ & $VFID_{RexNeXt}$$\downarrow$ & $VFID_{I3D}$$\downarrow$ & $VFID_{RexNeXt}$$\downarrow$\\
    \hline
    
    ViViD~\cite{vivid} & \textbf{0.8889} & \underline{0.0876} & \textbf{10.2367} & \underline{0.1785} & 16.4684 & 0.6807\\
    CatV$^2$TON~\cite{catv2ton} & 0.8670 & 0.1144 & 12.1280 & 0.1798 & 16.8880 & \textbf{0.3454}\\
    3DV-TON (Ours) & \underline{0.8801} & \textbf{0.0857} & \underline{10.7682} & \textbf{0.1420} & \textbf{14.5499} & \underline{0.4217}\\
    \hline
    \end{tabular}}
    \caption{\textbf{Quantitative comparison on HR-VVT benchmark. } Best results are highlighted in bold, the second are underlined. }
    \label{tab:hr}
\end{table}

\begin{table}[tbp]
\centering
    \resizebox{\linewidth}{!}{
    \begin{tabular}{l|l|ccc}
    \hline
    Datasets & Method & Fidelity (\%) & Consistency (\%) & Overall Quality (\%) \\
    \hline
    
    \multirow{3}{*}{ViViD} & ViViD~\cite{vivid} & 24.55 & 20.25 & 20.39 \\
    &CatV$^2$TON~\cite{catv2ton} & 12.09  & 10.92 & 10.53\\
    &3DV-TON (Ours) & \textbf{63.36} & \textbf{68.83} & \textbf{69.08}\\

    \hline
    \multirow{3}{*}{HR-VVT} & ViViD~\cite{vivid} & 14.02 & 11.82 & 11.77\\
    &CatV$^2$TON~\cite{catv2ton} & 5.97 & 3.36 & 2.70 \\
    &3DV-TON (Ours) & \textbf{80.01} & \textbf{84.82} & \textbf{85.53}\\
    
    \hline
    \end{tabular}}
    \caption{\textbf{User preference rate on the HR-VVT benchmark and ViViD dataset.}}
    \label{tab:user}
\end{table}

\subsection{Quantitative Results} 

\noindent \textbf{Comparisons on ViViD dataset.} We report quantitative results with SSIM~\cite{ssim}, LPIPS~\cite{lpips} to evaluate the image visual quality in the paired setting, and use Video Frechet Inception Distance(VFID)~\cite{fid, fwgan} to measure the generation quality and temporal consistency in the both paired and unpaired setting, following~\cite{fwgan, clothformer, vivid, catv2ton}. VFID extracts features of video clips for computation using pre-trained video backbone I3D~\cite{i3d} and 3D-ResNeXt101~\cite{3dresnext}.
Our method employs a rectangular mask strategy that enlarges the area to be generated, which creates an unfair comparison. Nonetheless, as reported in Table~\ref{tab:vvt}, our method still achieves comparable results in SSIM and LPIPS metrics, while surpassing existing methods in the VFID metric. When we use the mask from ViViD~\cite{vivid}, our method delivers better results across all metrics.

\noindent \textbf{Comparisons on HR-VVT benchmark.} 
We compare the current state-of-the-art and code released video try-on method, ViViD~\cite{vivid} and CatV$^2$TON on our benchmark. 
As shown in Table~\ref{tab:hr}, although we use the larger mask, our method outperforms other works. This improvement can be attributed to the consistent texture features brought by our textured 3D guidance. We also demonstrated advantages in LPIPS, which proves that our method is capable of generating try-on results with better visual quality.

\noindent \textbf{User Study.} Considering that the current quantitative metrics are difficult to accurately evaluate the quality of the model in terms of human preference in the unpaired setting without ground truth. We conduct a user study that includes 130 video results and involved 20 annotators to provide a comprehensive comparison in terms of visual quality and motion consistency. Table~\ref{tab:user} shows that our 3DV-TON achieves better motion coherence and effectively restores clothing details (\textit{i.e.} ``Fidelity''), resulting in superior visual quality.

\begin{figure}[tbp]
    \centering
    \tabcolsep=0.02in
    \begin{tabular}{c}
    \includegraphics[width=3.18in]{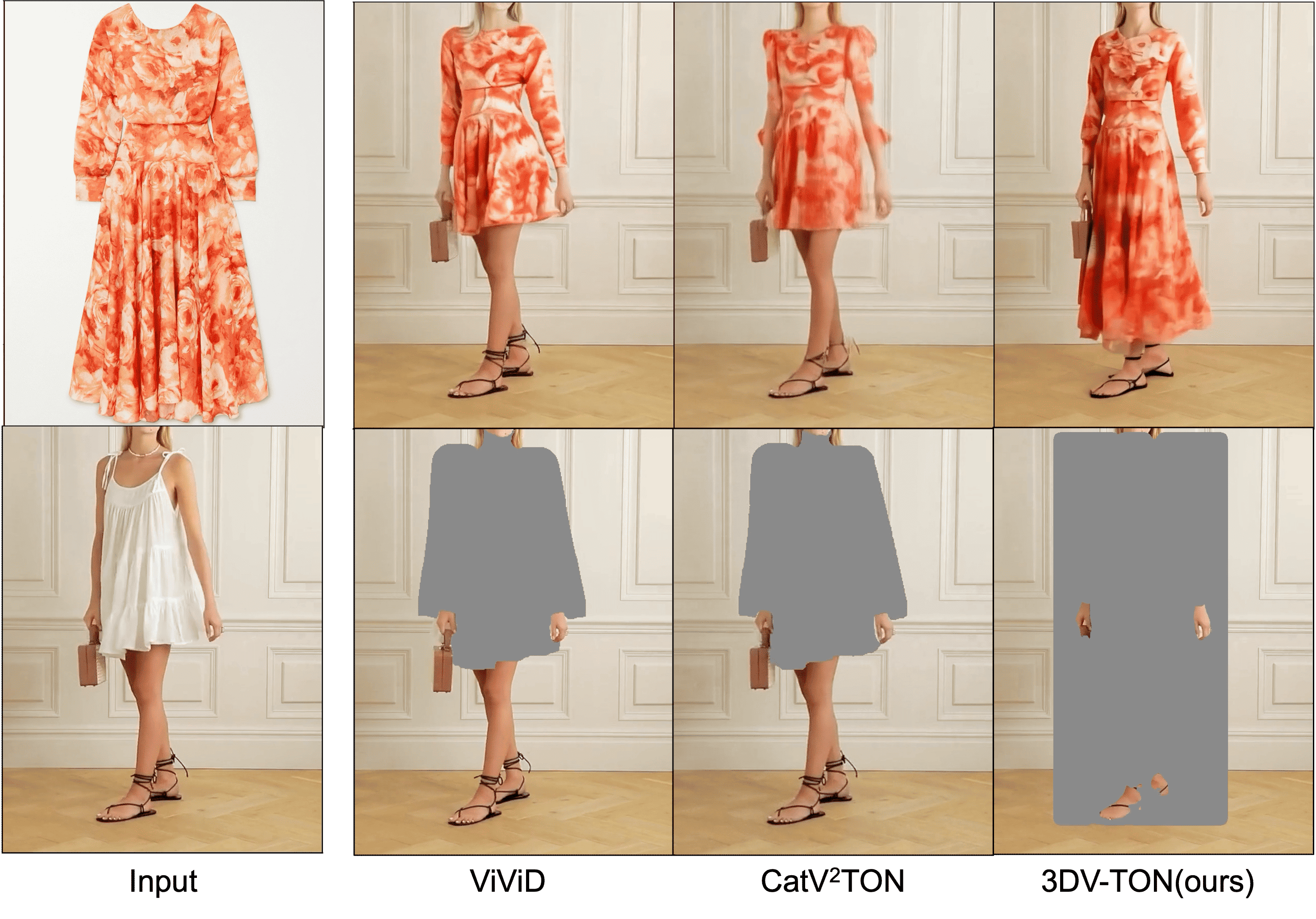}
    \end{tabular}
    \caption{\textbf{Ablations for the mask strategy.}}
    \label{fig:ab4_mask}
\end{figure}

\begin{figure}[tbp]
    \centering
    \tabcolsep=0.02in
    \begin{tabular}{c}
    \includegraphics[width=3.18in]{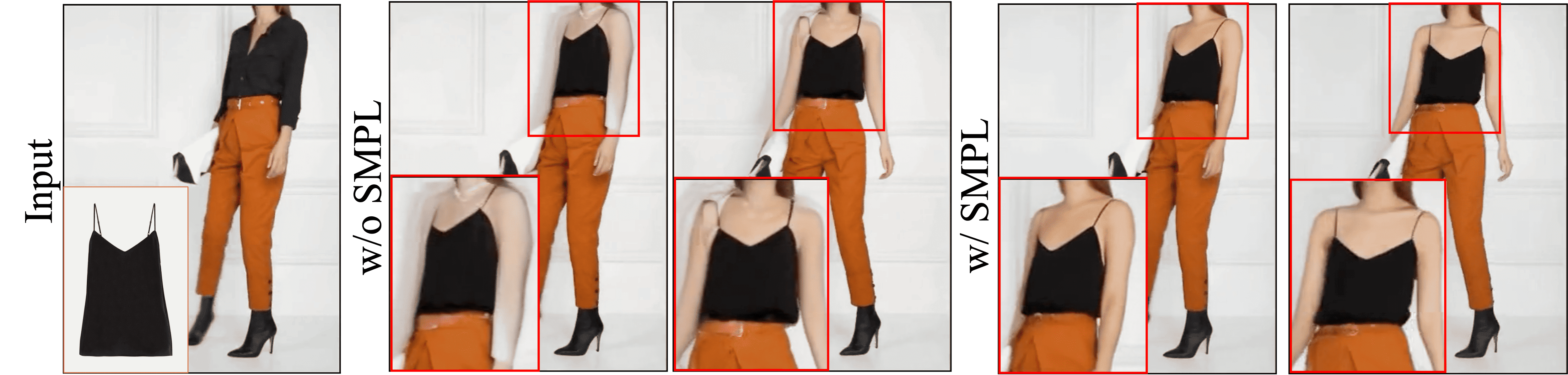}
    \end{tabular}
    \caption{\textbf{Ablations for the SMPL guidance.}}
    \label{fig:ab1_geo}
\end{figure}

\begin{figure}[tbp]
    \centering
    \tabcolsep=0.02in
    \begin{tabular}{c}
    \includegraphics[width=3in]{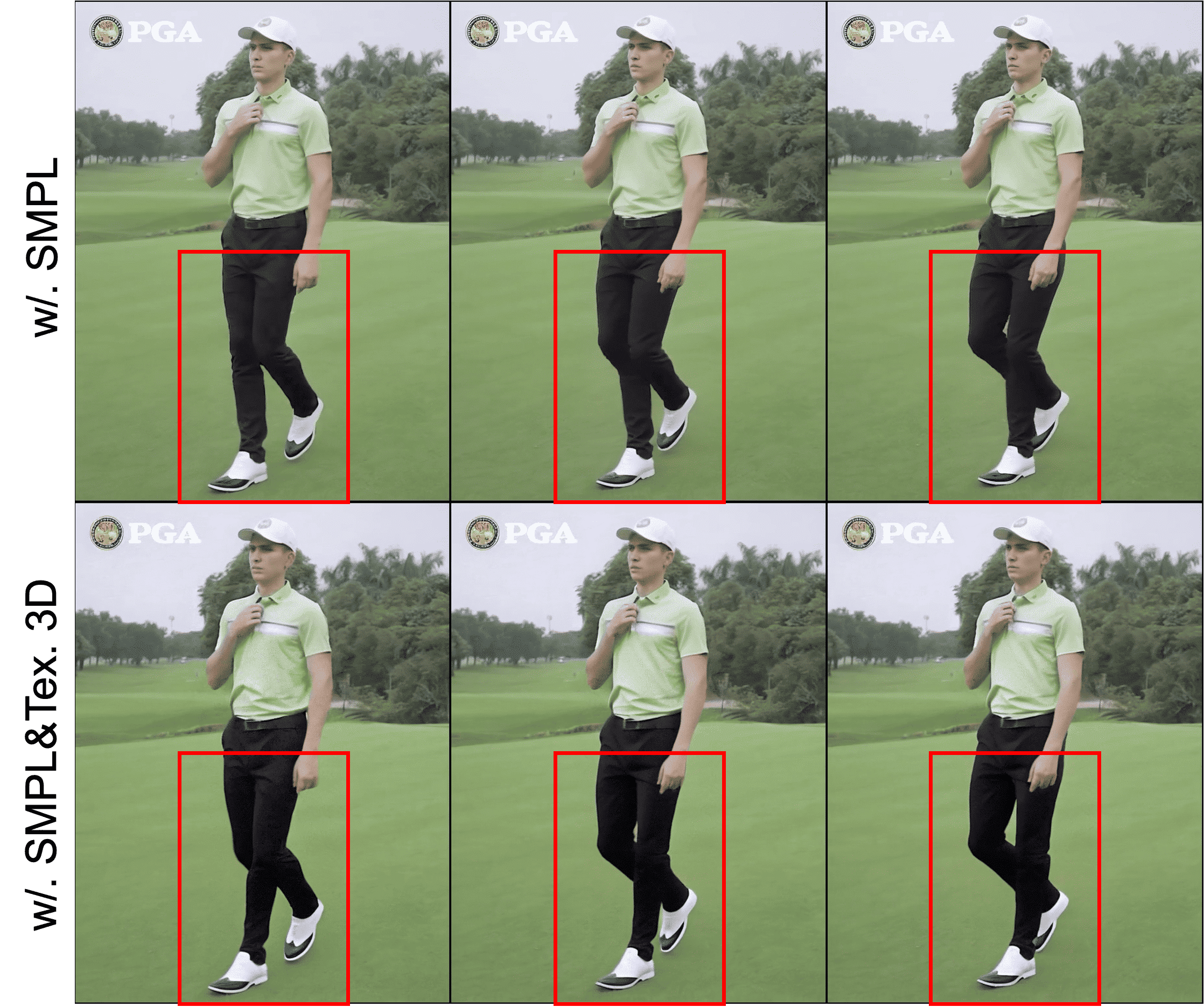}
    \end{tabular}
    \caption{\textbf{Ablations for the textured 3D guidance.} Textured 3D guidance helps to improve the motion conherence.}
    \label{fig:ab3_3d}
\end{figure}

\begin{figure}[tbp]
    \centering
    \tabcolsep=0.02in
    \begin{tabular}{c}
    \includegraphics[width=3.18in]{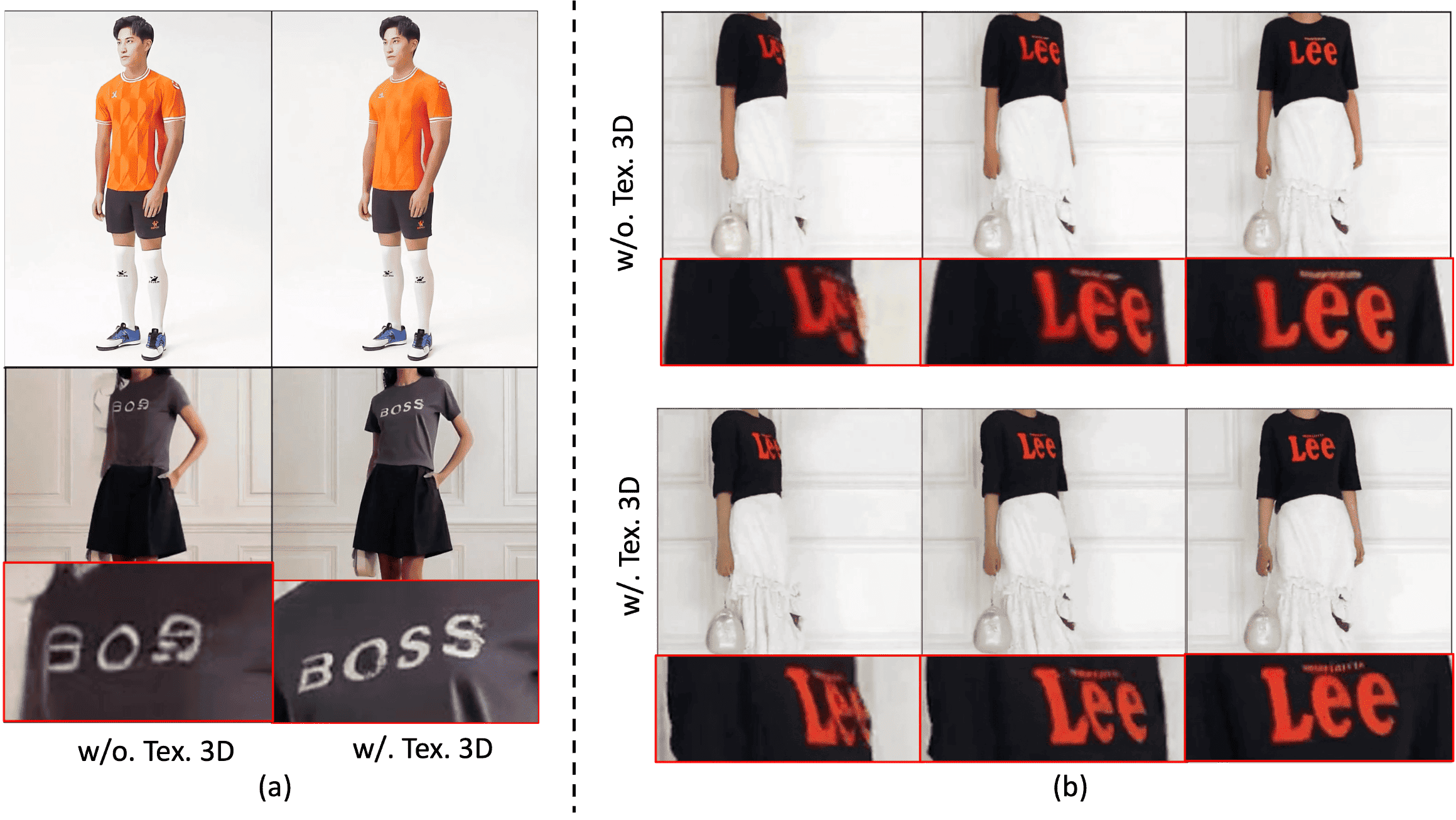}
    \end{tabular}
    \caption{\textbf{Ablations for our textured 3D guidance.} Textured 3D guidance helps to improve the clothing consistency.}
    \label{fig:ab2_3d}
\end{figure}

\subsection{Ablation Study}
\label{sec:ab}

We conducted ablation studies to verify the effectiveness of our textured 3D guidance. 

\noindent\textbf{Speed Analysis.} After optimizing the SMPL fitting process, our method is capable of completing the reconstruction in $\sim$30s, and generating a 32-frame video with diffusion after removing cross attention takes $\sim$35s under $768\times576$ resolution. Due to our use of single-image reconstruction, the diffusion model accounts for the majority of the inference time for longer videos.

\noindent \textbf{Mask Strategy.} Our robust rectangular mask strategy can effectively addresses the issue of try-on failures caused by the leakage of original clothing information in videos. 
Figure~\ref{fig:ab4_mask} demonstrates that our method can generate try-on results that align more closely with the target garment patterns.

\begin{table}[tbp]
\centering
    \resizebox{\linewidth}{!}{
    \begin{tabular}{cc|cccc}
    \hline
    SMPL & Tex. 3D & SSIM$\uparrow$ & LPIPS$\downarrow$ & $VFID_{I3D}$$\downarrow$ & $VFID_{RexNeXt}$$\downarrow$\\
    \hline
    
    && 0.858 & 0.078 & 5.236 & 1.0257\\
     \checkmark& & 0.880 & 0.059 & 4.087 & 0.5854\\
    \checkmark & \checkmark & \textbf{0.909} & \textbf{0.048} & \textbf{2.381} & \textbf{0.3011}\\
    \hline
    \end{tabular}}
    \caption{\textbf{Quantitative ablations for the 3D guidance.}}
   
    \label{tab:ab3}
\end{table}

\noindent \textbf{SMPL Guidance.} As shown in Figure~\ref{fig:ab1_geo}, the introduction of SMPL guidance helps in generating more accurate human bodies and properly fit clothing on the body. The person's arms and shoulders are accurately generated after using SMPL.

\noindent \textbf{Textured 3D Guidance.} Recent studies~\cite{videojam} demonstrate that conventional pixel reconstruction objective biases diffusion models toward appearance fidelity while compromising geometric accuracy, leading to motion artifacts.
As demonstrated in Figure~\ref{fig:ab3_3d}, while SMPL-based geometric guidance improves body structure estimation in masked regions, it exhibits persistent limb ambiguity during leg-crossing scenarios. Our textured 3D guidance resolves this limitation by supplementing explicit appearance constraints, effectively balancing visual quality and motion coherence.
Our texture 3D guidance ensures accurate clothing texture preservation across arbitrary poses and viewpoints. As shown in Figure~\ref{fig:ab2_3d} (a), our method faithfully reconstructs the ``boss'' logo while maintaining anatomically consistent body proportions during lateral rotation. Figure~\ref{fig:ab2_3d} (b) demonstrates viewpoint-consistent rendering of the ``lee'' text across dynamic poses. In Table~\ref{tab:ab3}, we present quantitative ablation experiments, where geometric features and textured 3D guidance significantly improved the SSIM, LPIPS, and VFID metrics.

\section{Conclusion}
In this paper, we propose 3DV-TON, a novel diffusion-based framework guided by geometric and textured 3D guidance. 
By leveraging SMPL as parametric body geometry and employing single-image reconstructed 3D humans as animatable textured 3D guidance to provide frame-specific appearance conditions,
3DV-TON alleviates the critical limitation of inconsistent results caused by existing methods' over-focus on appearance fidelity. The framework learns a geometrically plausible human body across diverse poses and viewpoints, while maintaining temporally consistent motion of clothing textures.
Quantitative and qualitative evaluations on existing datasets and our newly introduced HR-VVT demonstrate state-of-the-art performance in the video try-on task.

{
    \small
    \bibliographystyle{ieeenat_fullname}
    \bibliography{main}

\begin{thebibliography}{72}
\providecommand{\natexlab}[1]{#1}
\providecommand{\url}[1]{\texttt{#1}}
\expandafter\ifx\csname urlstyle\endcsname\relax
  \providecommand{\doi}[1]{doi: #1}\else
  \providecommand{\doi}{doi: \begingroup \urlstyle{rm}\Url}\fi

\bibitem[Bookstein and Green(1993)]{tps}
Fred~L Bookstein and WDK Green.
\newblock A thin-plate spline and the decomposition of deformations.
\newblock \emph{Mathematical Methods in Medical Imaging}, 2\penalty0 (14-28):\penalty0 3, 1993.

\bibitem[Cao et~al.(2022)Cao, Santo, Shi, Okura, and Matsushita]{bini}
Xu Cao, Hiroaki Santo, Boxin Shi, Fumio Okura, and Yasuyuki Matsushita.
\newblock Bilateral normal integration.
\newblock In \emph{European Conference on Computer Vision}, pages 552--567. Springer, 2022.

\bibitem[Carreira and Zisserman(2017)]{i3d}
Joao Carreira and Andrew Zisserman.
\newblock Quo vadis, action recognition? a new model and the kinetics dataset.
\newblock In \emph{proceedings of the IEEE Conference on Computer Vision and Pattern Recognition}, pages 6299--6308, 2017.

\bibitem[Chefer et~al.(2025)Chefer, Singer, Zohar, Kirstain, Polyak, Taigman, Wolf, and Sheynin]{videojam}
Hila Chefer, Uriel Singer, Amit Zohar, Yuval Kirstain, Adam Polyak, Yaniv Taigman, Lior Wolf, and Shelly Sheynin.
\newblock Videojam: Joint appearance-motion representations for enhanced motion generation in video models.
\newblock \emph{arXiv preprint arXiv:2502.02492}, 2025.

\bibitem[Choi et~al.(2021)Choi, Park, Lee, and Choo]{vitonhd}
Seunghwan Choi, Sunghyun Park, Minsoo Lee, and Jaegul Choo.
\newblock Viton-hd: High-resolution virtual try-on via misalignment-aware normalization.
\newblock In \emph{Proceedings of the IEEE/CVF conference on computer vision and pattern recognition}, pages 14131--14140, 2021.

\bibitem[Choi et~al.(2024)Choi, Kwak, Lee, Choi, and Shin]{idmvton}
Yisol Choi, Sangkyung Kwak, Kyungmin Lee, Hyungwon Choi, and Jinwoo Shin.
\newblock Improving diffusion models for virtual try-on.
\newblock \emph{arXiv preprint arXiv:2403.05139}, 2024.

\bibitem[Chong et~al.(2024)Chong, Dong, Li, Zhang, Zhang, Zhang, Zhao, and Liang]{catvton}
Zheng Chong, Xiao Dong, Haoxiang Li, Shiyue Zhang, Wenqing Zhang, Xujie Zhang, Hanqing Zhao, and Xiaodan Liang.
\newblock Catvton: Concatenation is all you need for virtual try-on with diffusion models.
\newblock \emph{arXiv preprint arXiv:2407.15886}, 2024.

\bibitem[Chong et~al.(2025)Chong, Zhang, Zhang, Zheng, Dong, Li, Wu, Jiang, and Liang]{catv2ton}
Zheng Chong, Wenqing Zhang, Shiyue Zhang, Jun Zheng, Xiao Dong, Haoxiang Li, Yiling Wu, Dongmei Jiang, and Xiaodan Liang.
\newblock Catv2ton: Taming diffusion transformers for vision-based virtual try-on with temporal concatenation.
\newblock \emph{arXiv preprint arXiv:2501.11325}, 2025.

\bibitem[Davide et~al.(2022)Davide, Matteo, Marcella, Federico, Fabio, and Rita]{dresscode}
Morelli Davide, Fincato Matteo, Cornia Marcella, Landi Federico, Cesari Fabio, and Cucchiara Rita.
\newblock Dress code: High-resolution multi-category virtual try-on.
\newblock In \emph{Proceedings of the IEEE/CVF Conference on Computer Vision and Pattern Recognition}, 2022.

\bibitem[De~Luigi et~al.(2023)De~Luigi, Li, Guillard, Salzmann, and Fua]{de2023drapenet}
Luca De~Luigi, Ren Li, Benoit Guillard, Mathieu Salzmann, and Pascal Fua.
\newblock Drapenet: Garment generation and self-supervised draping.
\newblock In \emph{Proceedings of the IEEE/CVF Conference on Computer Vision and Pattern Recognition}, pages 1451--1460, 2023.

\bibitem[Dong et~al.(2019)Dong, Liang, Shen, Wu, Chen, and Yin]{fwgan}
Haoye Dong, Xiaodan Liang, Xiaohui Shen, Bowen Wu, Bing-Cheng Chen, and Jian Yin.
\newblock Fw-gan: Flow-navigated warping gan for video virtual try-on.
\newblock In \emph{Proceedings of the IEEE/CVF international conference on computer vision}, pages 1161--1170, 2019.

\bibitem[Dong et~al.(2020)Dong, Liang, Zhang, Zhang, Shen, Xie, Wu, and Yin]{dong2020fashion}
Haoye Dong, Xiaodan Liang, Yixuan Zhang, Xujie Zhang, Xiaohui Shen, Zhenyu Xie, Bowen Wu, and Jian Yin.
\newblock Fashion editing with adversarial parsing learning.
\newblock In \emph{Proceedings of the IEEE/CVF conference on computer vision and pattern recognition}, pages 8120--8128, 2020.

\bibitem[Fang et~al.(2024)Fang, Zhai, Su, Song, Zhu, Wang, Chen, Liu, Cao, and Zha]{vivid}
Zixun Fang, Wei Zhai, Aimin Su, Hongliang Song, Kai Zhu, Mao Wang, Yu Chen, Zhiheng Liu, Yang Cao, and Zheng-Jun Zha.
\newblock Vivid: Video virtual try-on using diffusion models.
\newblock \emph{arXiv preprint arXiv:2405.11794}, 2024.

\bibitem[Feng et~al.(2021)Feng, Choutas, Bolkart, Tzionas, and Black]{pixie}
Yao Feng, Vasileios Choutas, Timo Bolkart, Dimitrios Tzionas, and Michael~J Black.
\newblock Collaborative regression of expressive bodies using moderation.
\newblock In \emph{2021 International Conference on 3D Vision (3DV)}, pages 792--804. IEEE, 2021.

\bibitem[Gao et~al.(2023)Gao, Chen, Zhang, Wang, Sun, Zhang, Bo, and Huang]{gao2023cloth2tex}
Daiheng Gao, Xu Chen, Xindi Zhang, Qi Wang, Ke Sun, Bang Zhang, Liefeng Bo, and Qixing Huang.
\newblock Cloth2tex: A customized cloth texture generation pipeline for 3d virtual try-on.
\newblock \emph{arXiv preprint arXiv:2308.04288}, 2023.

\bibitem[Ge et~al.(2021)Ge, Song, Zhang, Ge, Liu, and Luo]{ge2021parser}
Yuying Ge, Yibing Song, Ruimao Zhang, Chongjian Ge, Wei Liu, and Ping Luo.
\newblock Parser-free virtual try-on via distilling appearance flows.
\newblock In \emph{Proceedings of the IEEE/CVF conference on computer vision and pattern recognition}, pages 8485--8493, 2021.

\bibitem[Goodfellow et~al.(2020)Goodfellow, Pouget-Abadie, Mirza, Xu, Warde-Farley, Ozair, Courville, and Bengio]{gan}
Ian Goodfellow, Jean Pouget-Abadie, Mehdi Mirza, Bing Xu, David Warde-Farley, Sherjil Ozair, Aaron Courville, and Yoshua Bengio.
\newblock Generative adversarial networks.
\newblock \emph{Communications of the ACM}, 63\penalty0 (11):\penalty0 139--144, 2020.

\bibitem[Gou et~al.(2023)Gou, Sun, Zhang, Si, Qian, and Zhang]{dcivton}
Junhong Gou, Siyu Sun, Jianfu Zhang, Jianlou Si, Chen Qian, and Liqing Zhang.
\newblock Taming the power of diffusion models for high-quality virtual try-on with appearance flow.
\newblock In \emph{Proceedings of the 31st ACM International Conference on Multimedia}, 2023.

\bibitem[G{\"u}ler et~al.(2018)G{\"u}ler, Neverova, and Kokkinos]{guler2018densepose}
R{\i}za~Alp G{\"u}ler, Natalia Neverova, and Iasonas Kokkinos.
\newblock Densepose: Dense human pose estimation in the wild.
\newblock In \emph{Proceedings of the IEEE conference on computer vision and pattern recognition}, pages 7297--7306, 2018.

\bibitem[Guo et~al.(2024)Guo, Yang, Rao, Liang, Wang, Qiao, Agrawala, Lin, and Dai]{animatediff}
Yuwei Guo, Ceyuan Yang, Anyi Rao, Zhengyang Liang, Yaohui Wang, Yu Qiao, Maneesh Agrawala, Dahua Lin, and Bo Dai.
\newblock Animatediff: Animate your personalized text-to-image diffusion models without specific tuning.
\newblock \emph{International Conference on Learning Representations}, 2024.

\bibitem[Han et~al.(2018)Han, Wu, Wu, Yu, and Davis]{viton}
Xintong Han, Zuxuan Wu, Zhe Wu, Ruichi Yu, and Larry~S Davis.
\newblock Viton: An image-based virtual try-on network.
\newblock In \emph{Proceedings of the IEEE conference on computer vision and pattern recognition}, pages 7543--7552, 2018.

\bibitem[Han et~al.(2019)Han, Hu, Huang, and Scott]{han2019clothflow}
Xintong Han, Xiaojun Hu, Weilin Huang, and Matthew~R Scott.
\newblock Clothflow: A flow-based model for clothed person generation.
\newblock In \emph{Proceedings of the IEEE/CVF international conference on computer vision}, pages 10471--10480, 2019.

\bibitem[Hara et~al.(2018)Hara, Kataoka, and Satoh]{3dresnext}
Kensho Hara, Hirokatsu Kataoka, and Yutaka Satoh.
\newblock Can spatiotemporal 3d cnns retrace the history of 2d cnns and imagenet?
\newblock In \emph{Proceedings of the IEEE conference on Computer Vision and Pattern Recognition}, pages 6546--6555, 2018.

\bibitem[He et~al.(2022)He, Song, and Xiang]{he2022style}
Sen He, Yi-Zhe Song, and Tao Xiang.
\newblock Style-based global appearance flow for virtual try-on.
\newblock In \emph{Proceedings of the IEEE/CVF Conference on Computer Vision and Pattern Recognition}, pages 3470--3479, 2022.

\bibitem[He et~al.(2024)He, Chen, Wang, Li, Torr, and Lin]{he2024wildvidfit}
Zijian He, Peixin Chen, Guangrun Wang, Guanbin Li, Philip~HS Torr, and Liang Lin.
\newblock Wildvidfit: Video virtual try-on in the wild via image-based controlled diffusion models.
\newblock In \emph{European Conference on Computer Vision}, pages 123--139. Springer, 2024.

\bibitem[Ho and Salimans(2021)]{cfg}
Jonathan Ho and Tim Salimans.
\newblock Classifier-free diffusion guidance.
\newblock In \emph{NeurIPS 2021 Workshop on Deep Generative Models and Downstream Applications}, 2021.

\bibitem[Ho et~al.(2020)Ho, Jain, and Abbeel]{ddpm}
Jonathan Ho, Ajay Jain, and Pieter Abbeel.
\newblock Denoising diffusion probabilistic models.
\newblock \emph{Advances in neural information processing systems}, 33:\penalty0 6840--6851, 2020.

\bibitem[Huang et~al.(2023)Huang, Gojcic, Atzmon, Litany, Fidler, and Williams]{nksr}
Jiahui Huang, Zan Gojcic, Matan Atzmon, Or Litany, Sanja Fidler, and Francis Williams.
\newblock Neural kernel surface reconstruction.
\newblock In \emph{Proceedings of the IEEE/CVF Conference on Computer Vision and Pattern Recognition}, pages 4369--4379, 2023.

\bibitem[Huang et~al.(2022)Huang, Li, Xie, Kampffmeyer, Liang, et~al.]{huang2022towards}
Zaiyu Huang, Hanhui Li, Zhenyu Xie, Michael Kampffmeyer, Xiaodan Liang, et~al.
\newblock Towards hard-pose virtual try-on via 3d-aware global correspondence learning.
\newblock \emph{Advances in Neural Information Processing Systems}, 35:\penalty0 32736--32748, 2022.

\bibitem[Jiang et~al.(2022)Jiang, Wang, Yan, and Liu]{clothformer}
Jianbin Jiang, Tan Wang, He Yan, and Junhui Liu.
\newblock Clothformer: Taming video virtual try-on in all module.
\newblock In \emph{Proceedings of the IEEE/CVF Conference on Computer Vision and Pattern Recognition}, pages 10799--10808, 2022.

\bibitem[Kazhdan et~al.(2006)Kazhdan, Bolitho, and Hoppe]{psr}
Michael Kazhdan, Matthew Bolitho, and Hugues Hoppe.
\newblock Poisson surface reconstruction.
\newblock In \emph{Proceedings of the fourth Eurographics symposium on Geometry processing}, 2006.

\bibitem[Khirodkar et~al.(2025)Khirodkar, Bagautdinov, Martinez, Zhaoen, James, Selednik, Anderson, and Saito]{sapiens}
Rawal Khirodkar, Timur Bagautdinov, Julieta Martinez, Su Zhaoen, Austin James, Peter Selednik, Stuart Anderson, and Shunsuke Saito.
\newblock Sapiens: Foundation for human vision models.
\newblock In \emph{European Conference on Computer Vision}, pages 206--228. Springer, 2025.

\bibitem[Kim et~al.(2024)Kim, Gu, Park, Park, and Choo]{stablevton}
Jeongho Kim, Guojung Gu, Minho Park, Sunghyun Park, and Jaegul Choo.
\newblock Stableviton: Learning semantic correspondence with latent diffusion model for virtual try-on.
\newblock In \emph{Proceedings of the IEEE/CVF Conference on Computer Vision and Pattern Recognition}, pages 8176--8185, 2024.

\bibitem[Kingma(2013)]{vae}
Diederik~P Kingma.
\newblock Auto-encoding variational bayes.
\newblock \emph{arXiv preprint arXiv:1312.6114}, 2013.

\bibitem[Kirillov et~al.(2023)Kirillov, Mintun, Ravi, Mao, Rolland, Gustafson, Xiao, Whitehead, Berg, Lo, et~al.]{kirillov2023segment}
Alexander Kirillov, Eric Mintun, Nikhila Ravi, Hanzi Mao, Chloe Rolland, Laura Gustafson, Tete Xiao, Spencer Whitehead, Alexander~C Berg, Wan-Yen Lo, et~al.
\newblock Segment anything.
\newblock In \emph{Proceedings of the IEEE/CVF international conference on computer vision}, pages 4015--4026, 2023.

\bibitem[Li et~al.(2021)Li, Xu, Chen, Bian, Yang, and Lu]{hybrik}
Jiefeng Li, Chao Xu, Zhicun Chen, Siyuan Bian, Lixin Yang, and Cewu Lu.
\newblock Hybrik: A hybrid analytical-neural inverse kinematics solution for 3d human pose and shape estimation.
\newblock In \emph{Proceedings of the IEEE/CVF Conference on Computer Vision and Pattern Recognition}, pages 3383--3393, 2021.

\bibitem[Li et~al.(2023)Li, Bian, Xu, Chen, Yang, and Lu]{hybrikx}
Jiefeng Li, Siyuan Bian, Chao Xu, Zhicun Chen, Lixin Yang, and Cewu Lu.
\newblock Hybrik-x: Hybrid analytical-neural inverse kinematics for whole-body mesh recovery.
\newblock \emph{arXiv preprint arXiv:2304.05690}, 2023.

\bibitem[Li et~al.(2020)Li, Xu, Wei, and Yang]{li2020self}
Peike Li, Yunqiu Xu, Yunchao Wei, and Yi Yang.
\newblock Self-correction for human parsing.
\newblock \emph{IEEE Transactions on Pattern Analysis and Machine Intelligence}, 44\penalty0 (6):\penalty0 3260--3271, 2020.

\bibitem[Li et~al.(2024)Li, Chen, Larionov, Sarafianos, Matusik, and Stuyck]{diffavatar}
Yifei Li, Hsiao-yu Chen, Egor Larionov, Nikolaos Sarafianos, Wojciech Matusik, and Tuur Stuyck.
\newblock Diffavatar: Simulation-ready garment optimization with differentiable simulation.
\newblock In \emph{Proceedings of the IEEE/CVF Conference on Computer Vision and Pattern Recognition}, pages 4368--4378, 2024.

\bibitem[Lin et~al.(2024)Lin, Liu, Li, and Yang]{vpred}
Shanchuan Lin, Bingchen Liu, Jiashi Li, and Xiao Yang.
\newblock Common diffusion noise schedules and sample steps are flawed.
\newblock In \emph{Proceedings of the IEEE/CVF winter conference on applications of computer vision}, pages 5404--5411, 2024.

\bibitem[Loper et~al.(2023)Loper, Mahmood, Romero, Pons-Moll, and Black]{smpl}
Matthew Loper, Naureen Mahmood, Javier Romero, Gerard Pons-Moll, and Michael~J Black.
\newblock Smpl: A skinned multi-person linear model.
\newblock In \emph{Seminal Graphics Papers: Pushing the Boundaries, Volume 2}, pages 851--866. 2023.

\bibitem[Martin et~al.(2017)Martin, Hubert, Thomas, Bernhard, and Sepp]{fid}
Heusel Martin, Ramsauer Hubert, Unterthiner Thomas, Nessler Bernhard, and Hochreiter Sepp.
\newblock Gans trained by a two time-scale update rule converge to a local nash equilibrium.
\newblock \emph{Advances in neural information processing systems}, 30:\penalty0 6626--6637, 2017.

\bibitem[Minar and Ahn(2020)]{minar2020cloth}
Matiur~Rahman Minar and Heejune Ahn.
\newblock Cloth-vton: Clothing three-dimensional reconstruction for hybrid image-based virtual try-on.
\newblock In \emph{Proceedings of the Asian conference on computer vision}, 2020.

\bibitem[Morelli et~al.(2023)Morelli, Baldrati, Cartella, Cornia, Bertini, and Cucchiara]{ladivton}
Davide Morelli, Alberto Baldrati, Giuseppe Cartella, Marcella Cornia, Marco Bertini, and Rita Cucchiara.
\newblock Ladi-vton: Latent diffusion textual-inversion enhanced virtual try-on.
\newblock In \emph{Proceedings of the 31st ACM International Conference on Multimedia}, pages 8580--8589, 2023.

\bibitem[Pavlakos et~al.(2019)Pavlakos, Choutas, Ghorbani, Bolkart, Osman, Tzionas, and Black]{smplx}
Georgios Pavlakos, Vasileios Choutas, Nima Ghorbani, Timo Bolkart, Ahmed A.~A. Osman, Dimitrios Tzionas, and Michael~J. Black.
\newblock Expressive body capture: 3d hands, face, and body from a single image.
\newblock In \emph{Proceedings IEEE Conf. on Computer Vision and Pattern Recognition (CVPR)}, 2019.

\bibitem[Peebles and Xie(2023)]{dit}
William Peebles and Saining Xie.
\newblock Scalable diffusion models with transformers.
\newblock In \emph{Proceedings of the IEEE/CVF international conference on computer vision}, pages 4195--4205, 2023.

\bibitem[Polyak et~al.(2024)Polyak, Zohar, Brown, Tjandra, Sinha, Lee, Vyas, Shi, Ma, Chuang, et~al.]{moviegen}
Adam Polyak, Amit Zohar, Andrew Brown, Andros Tjandra, Animesh Sinha, Ann Lee, Apoorv Vyas, Bowen Shi, Chih-Yao Ma, Ching-Yao Chuang, et~al.
\newblock Movie gen: A cast of media foundation models.
\newblock \emph{arXiv preprint arXiv:2410.13720}, 2024.

\bibitem[Qiu et~al.(2025)Qiu, Gu, Li, Zuo, Shen, Zhang, Qiu, Yuan, Chen, Dong, et~al.]{lhm}
Lingteng Qiu, Xiaodong Gu, Peihao Li, Qi Zuo, Weichao Shen, Junfei Zhang, Kejie Qiu, Weihao Yuan, Guanying Chen, Zilong Dong, et~al.
\newblock Lhm: Large animatable human reconstruction model from a single image in seconds.
\newblock \emph{arXiv preprint arXiv:2503.10625}, 2025.

\bibitem[Rombach et~al.(2022)Rombach, Blattmann, Lorenz, Esser, and Ommer]{sd}
Robin Rombach, Andreas Blattmann, Dominik Lorenz, Patrick Esser, and Bj{\"o}rn Ommer.
\newblock High-resolution image synthesis with latent diffusion models.
\newblock In \emph{Proceedings of the IEEE/CVF conference on computer vision and pattern recognition}, pages 10684--10695, 2022.

\bibitem[Ronneberger et~al.(2015)Ronneberger, Fischer, and Brox]{unet}
Olaf Ronneberger, Philipp Fischer, and Thomas Brox.
\newblock U-net: Convolutional networks for biomedical image segmentation.
\newblock In \emph{Medical image computing and computer-assisted intervention--MICCAI 2015: 18th international conference, Munich, Germany, October 5-9, 2015, proceedings, part III 18}, pages 234--241. Springer, 2015.

\bibitem[Shen et~al.(2024)Shen, Pi, Xia, Cen, Peng, Hu, Bao, Hu, and Zhou]{gvhmr}
Zehong Shen, Huaijin Pi, Yan Xia, Zhi Cen, Sida Peng, Zechen Hu, Hujun Bao, Ruizhen Hu, and Xiaowei Zhou.
\newblock World-grounded human motion recovery via gravity-view coordinates.
\newblock In \emph{SIGGRAPH Asia Conference Proceedings}, 2024.

\bibitem[Shin et~al.(2024)Shin, Kim, Halilaj, and Black]{wham}
Soyong Shin, Juyong Kim, Eni Halilaj, and Michael~J Black.
\newblock Wham: Reconstructing world-grounded humans with accurate 3d motion.
\newblock In \emph{Proceedings of the IEEE/CVF Conference on Computer Vision and Pattern Recognition}, pages 2070--2080, 2024.

\bibitem[Sumner et~al.(2007)Sumner, Schmid, and Pauly]{lbs}
Robert~W Sumner, Johannes Schmid, and Mark Pauly.
\newblock Embedded deformation for shape manipulation.
\newblock In \emph{ACM siggraph 2007 papers}, pages 80--es. 2007.

\bibitem[Wang et~al.(2018)Wang, Zheng, Liang, Chen, Lin, and Yang]{wang2018toward}
Bochao Wang, Huabin Zheng, Xiaodan Liang, Yimin Chen, Liang Lin, and Meng Yang.
\newblock Toward characteristic-preserving image-based virtual try-on network.
\newblock In \emph{Proceedings of the European conference on computer vision (ECCV)}, pages 589--604, 2018.

\bibitem[Wang et~al.(2024)Wang, Dai, Chan, Zhou, Zhang, and Liu]{wang2024gpd}
Yuanbin Wang, Weilun Dai, Long Chan, Huanyu Zhou, Aixi Zhang, and Si Liu.
\newblock Gpd-vvto: Preserving garment details in video virtual try-on.
\newblock In \emph{Proceedings of the 32nd ACM International Conference on Multimedia}, pages 7133--7142, 2024.

\bibitem[Wang et~al.(2004)Wang, Bovik, Sheikh, and Simoncelli]{ssim}
Zhou Wang, Alan~C Bovik, Hamid~R Sheikh, and Eero~P Simoncelli.
\newblock Image quality assessment: from error visibility to structural similarity.
\newblock \emph{IEEE transactions on image processing}, 13\penalty0 (4):\penalty0 600--612, 2004.

\bibitem[Weng et~al.(2024)Weng, Liu, Tan, Xu, Zhou, Yeung-Levy, and Yang]{weng2024template}
Zhenzhen Weng, Jingyuan Liu, Hao Tan, Zhan Xu, Yang Zhou, Serena Yeung-Levy, and Jimei Yang.
\newblock Template-free single-view 3d human digitalization with diffusion-guided lrm.
\newblock \emph{arXiv preprint arXiv:2401.12175}, 2024.

\bibitem[Xie et~al.(2023)Xie, Huang, Dong, Zhao, Dong, Zhang, Zhu, and Liang]{xie2023gp}
Zhenyu Xie, Zaiyu Huang, Xin Dong, Fuwei Zhao, Haoye Dong, Xijin Zhang, Feida Zhu, and Xiaodan Liang.
\newblock Gp-vton: Towards general purpose virtual try-on via collaborative local-flow global-parsing learning.
\newblock In \emph{Proceedings of the IEEE/CVF Conference on Computer Vision and Pattern Recognition}, pages 23550--23559, 2023.

\bibitem[Xiu et~al.(2022)Xiu, Yang, Tzionas, and Black]{icon}
Yuliang Xiu, Jinlong Yang, Dimitrios Tzionas, and Michael~J. Black.
\newblock {ICON}: {I}mplicit {C}lothed humans {O}btained from {N}ormals.
\newblock In \emph{Proceedings of the IEEE/CVF Conference on Computer Vision and Pattern Recognition (CVPR)}, pages 13296--13306, 2022.

\bibitem[Xiu et~al.(2023)Xiu, Yang, Cao, Tzionas, and Black]{econ}
Yuliang Xiu, Jinlong Yang, Xu Cao, Dimitrios Tzionas, and Michael~J. Black.
\newblock {ECON: Explicit Clothed humans Optimized via Normal integration}.
\newblock In \emph{Proceedings of the IEEE/CVF Conference on Computer Vision and Pattern Recognition (CVPR)}, 2023.

\bibitem[Xu et~al.(2024{\natexlab{a}})Xu, Gu, Chen, and Chen]{ootd}
Yuhao Xu, Tao Gu, Weifeng Chen, and Chengcai Chen.
\newblock Ootdiffusion: Outfitting fusion based latent diffusion for controllable virtual try-on.
\newblock \emph{arXiv preprint arXiv:2403.01779}, 2024{\natexlab{a}}.

\bibitem[Xu et~al.(2024{\natexlab{b}})Xu, Chen, Wang, Xing, Zhai, Sang, Lan, Xiao, and Gao]{tunneltryon}
Zhengze Xu, Mengting Chen, Zhao Wang, Linyu Xing, Zhonghua Zhai, Nong Sang, Jinsong Lan, Shuai Xiao, and Changxin Gao.
\newblock Tunnel try-on: Excavating spatial-temporal tunnels for high-quality virtual try-on in videos.
\newblock In \emph{Proceedings of the 32nd ACM International Conference on Multimedia}, pages 3199--3208, 2024{\natexlab{b}}.

\bibitem[Xu et~al.(2024{\natexlab{c}})Xu, Zhang, Liew, Yan, Liu, Zhang, Feng, and Shou]{magicanimate}
Zhongcong Xu, Jianfeng Zhang, Jun~Hao Liew, Hanshu Yan, Jia-Wei Liu, Chenxu Zhang, Jiashi Feng, and Mike~Zheng Shou.
\newblock Magicanimate: Temporally consistent human image animation using diffusion model.
\newblock In \emph{Proceedings of the IEEE/CVF Conference on Computer Vision and Pattern Recognition}, pages 1481--1490, 2024{\natexlab{c}}.

\bibitem[Zhang et~al.(2023)Zhang, Tian, Zhang, Li, An, Sun, and Liu]{pymaf}
Hongwen Zhang, Yating Tian, Yuxiang Zhang, Mengcheng Li, Liang An, Zhenan Sun, and Yebin Liu.
\newblock Pymaf-x: Towards well-aligned full-body model regression from monocular images.
\newblock \emph{IEEE Transactions on Pattern Analysis and Machine Intelligence}, 45\penalty0 (10):\penalty0 12287--12303, 2023.

\bibitem[Zhang et~al.(2021)Zhang, Li, Lai, and Yang]{zhang2021pise}
Jinsong Zhang, Kun Li, Yu-Kun Lai, and Jingyu Yang.
\newblock Pise: Person image synthesis and editing with decoupled gan.
\newblock In \emph{Proceedings of the IEEE/CVF Conference on Computer Vision and Pattern Recognition}, pages 7982--7990, 2021.

\bibitem[Zhang et~al.(2018)Zhang, Isola, Efros, Shechtman, and Wang]{lpips}
Richard Zhang, Phillip Isola, Alexei~A Efros, Eli Shechtman, and Oliver Wang.
\newblock The unreasonable effectiveness of deep features as a perceptual metric.
\newblock In \emph{Proceedings of the IEEE conference on computer vision and pattern recognition}, pages 586--595, 2018.

\bibitem[Zhang et~al.(2024)Zhang, Yang, and Yang]{sifu}
Zechuan Zhang, Zongxin Yang, and Yi Yang.
\newblock Sifu: Side-view conditioned implicit function for real-world usable clothed human reconstruction.
\newblock In \emph{Proceedings of the IEEE/CVF Conference on Computer Vision and Pattern Recognition (CVPR)}, pages 9936--9947, 2024.

\bibitem[Zhao et~al.(2021)Zhao, Xie, Kampffmeyer, Dong, Han, Zheng, Zhang, and Liang]{zhao2021m3d}
Fuwei Zhao, Zhenyu Xie, Michael Kampffmeyer, Haoye Dong, Songfang Han, Tianxiang Zheng, Tao Zhang, and Xiaodan Liang.
\newblock M3d-vton: A monocular-to-3d virtual try-on network.
\newblock In \emph{Proceedings of the IEEE/CVF International Conference on Computer Vision}, pages 13239--13249, 2021.

\bibitem[Zheng et~al.(2024)Zheng, Zhao, Xu, Dong, and Liang]{vitondit}
Jun Zheng, Fuwei Zhao, Youjiang Xu, Xin Dong, and Xiaodan Liang.
\newblock Viton-dit: Learning in-the-wild video try-on from human dance videos via diffusion transformers.
\newblock \emph{arXiv preprint arXiv:2405.18326}, 2024.

\bibitem[Zhong et~al.(2021)Zhong, Wu, Tan, Lin, and Wu]{mvton}
Xiaojing Zhong, Zhonghua Wu, Taizhe Tan, Guosheng Lin, and Qingyao Wu.
\newblock Mv-ton: Memory-based video virtual try-on network.
\newblock In \emph{Proceedings of the 29th ACM International Conference on Multimedia}, pages 908--916, 2021.

\bibitem[Zhou et~al.(2024)Zhou, Wang, Chen, Bai, Li, Zhang, Xu, Yang, and Wang]{realisdance}
Jingkai Zhou, Benzhi Wang, Weihua Chen, Jingqi Bai, Dongyang Li, Aixi Zhang, Hao Xu, Mingyang Yang, and Fan Wang.
\newblock Realisdance: Equip controllable character animation with realistic hands.
\newblock \emph{arXiv preprint arXiv:2409.06202}, 2024.

\bibitem[Zhu et~al.(2024)Zhu, Chen, Dai, Dong, Xu, Cao, Yao, Zhu, and Zhu]{champ}
Shenhao Zhu, Junming~Leo Chen, Zuozhuo Dai, Zilong Dong, Yinghui Xu, Xun Cao, Yao Yao, Hao Zhu, and Siyu Zhu.
\newblock Champ: Controllable and consistent human image animation with 3d parametric guidance.
\newblock In \emph{European Conference on Computer Vision}, pages 145--162. Springer, 2024.

\end{thebibliography}
}

\clearpage

\maketitlesupplementary

\appendix

\section{HR-VVT benchmark.} 
\begin{figure*}[h]
    \centering
    \tabcolsep=0.02in
    \begin{tabular}{c}
        \includegraphics[width=6in]{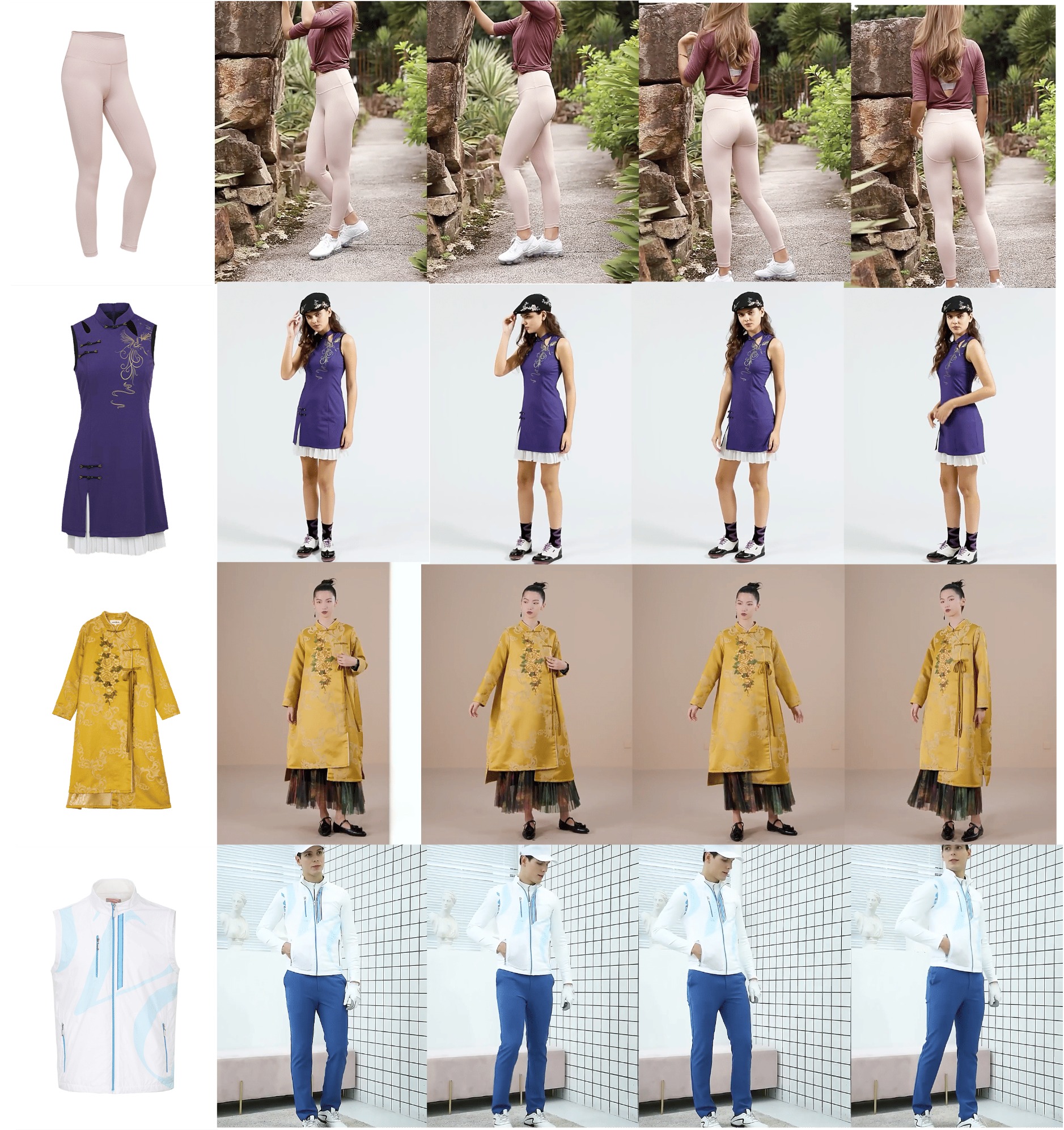}\\
    \end{tabular}
    \vspace{-.1in}
    \caption{\textbf{Illustration of the HR-VVT benckmark.}}
    \label{fig:dataset}
\end{figure*}

Owing to the limitations of the exist datasets, which exhibits relatively simple scenarios, it is challenging to accurately assess video try-on methods.
Therefore, we have constructed a high-resolution (\textasciitilde720p) video try-on benchmark called HR-VVT that includes 130 videos with 50 upper-body clothing, 40 lower-body clothing, and 40 dresses, with a variety of garments and motions in complex scenarios. Figure~\ref{fig:dataset} show some examples in our benchmark.

Our HR-VVT benchmark was sourced from e-commerce platforms for research purposes and will remain strictly reserved for academic use. Our framework contains no personal identity information, with facial regions excluded from inpainting operations to ensure privacy preservation during the training process.

\section{Disccusion}
\noindent\textbf{Limitation.}
Although we have significantly reduced the reconstruction time of clothed 3D humans by improving the optimization objectives of SMPL refinement and keeping it within an acceptable inference time, this is still insufficient in scenarios with higher speed requirements. Recently, works~\cite{lhm} on reconstructing animatable clothed 3D humans using a single feed-forward approach has greatly accelerated inference times and achieved remarkable improvements in visual quality. We believe that updating our 3D guidance pipeline to a single feed-forward paradigm can accelerate the reconstruction process, further advancing the application of textured 3D human guidance in more scenarios.

\noindent\textbf{Potential societal impact.}
This paper delves into the realm of video try-on generation. Because of the powerful generative capacity, these models pose risks such as the potential for misinformation and the creation of fake videos. We sincerely remind users to pay attention to generated content.
Besides, it is crucial to prioritize privacy and consent, as generative models frequently rely on vast datasets that may include sensitive information. Users must remain vigilant about these considerations to uphold ethical standards in their applications.
Note that our method only focus on technical aspect. Both videos and model weights used in this paper will be open-released.

\section{Animatable Textured 3D Guidance}
\noindent\textbf{Refine SMPL-X.} 
Since our clothed human reconstruction method is based on the SMPL-X~\cite{smpl,smplx} model, it is important to accurately align the estimated body and clothing silhouette. In practice, human pose and shape (HPS) regressors~\cite{hybrikx, hybrik, gvhmr} can not give pixel-aligned SMPL-X fits. We refine the SMPL-X parameters by minimizing $\mathcal{L}_{\text{SMPL-X}}$ in Section 3.1 of our paper.

Unlike the optimization of shape $\beta$, pose $\theta$, and translation $t$ of SMPL-X in ICON~\cite{icon}, Since we primarily focus on the clothing area and do not have high accuracy requirements for the body pose details, we adjust the optimization target without optimizing the SMPL pose $\theta$. This allows us to significantly reduce the number of optimization steps to reduce the reconstruction time (\textasciitilde 30s). And we optimize the shape $\beta$, camera scale $s$, and translation $t$ to mitigate the anomalies in loss caused by incomplete human body parts and inaccurate camera estimation in real data, which may lead to errors in the refined SMPL-X. We additionally introduce a unidirectional regularization penalty to prevent the incorrect decrease in loss caused by abnormal reduction in camera scale caused by partial bodies present in the training data.

\begin{figure*}[h]
    \centering
    \tabcolsep=0.02in
    \begin{tabular}{cccccc}
        \includegraphics[width=1.08in]{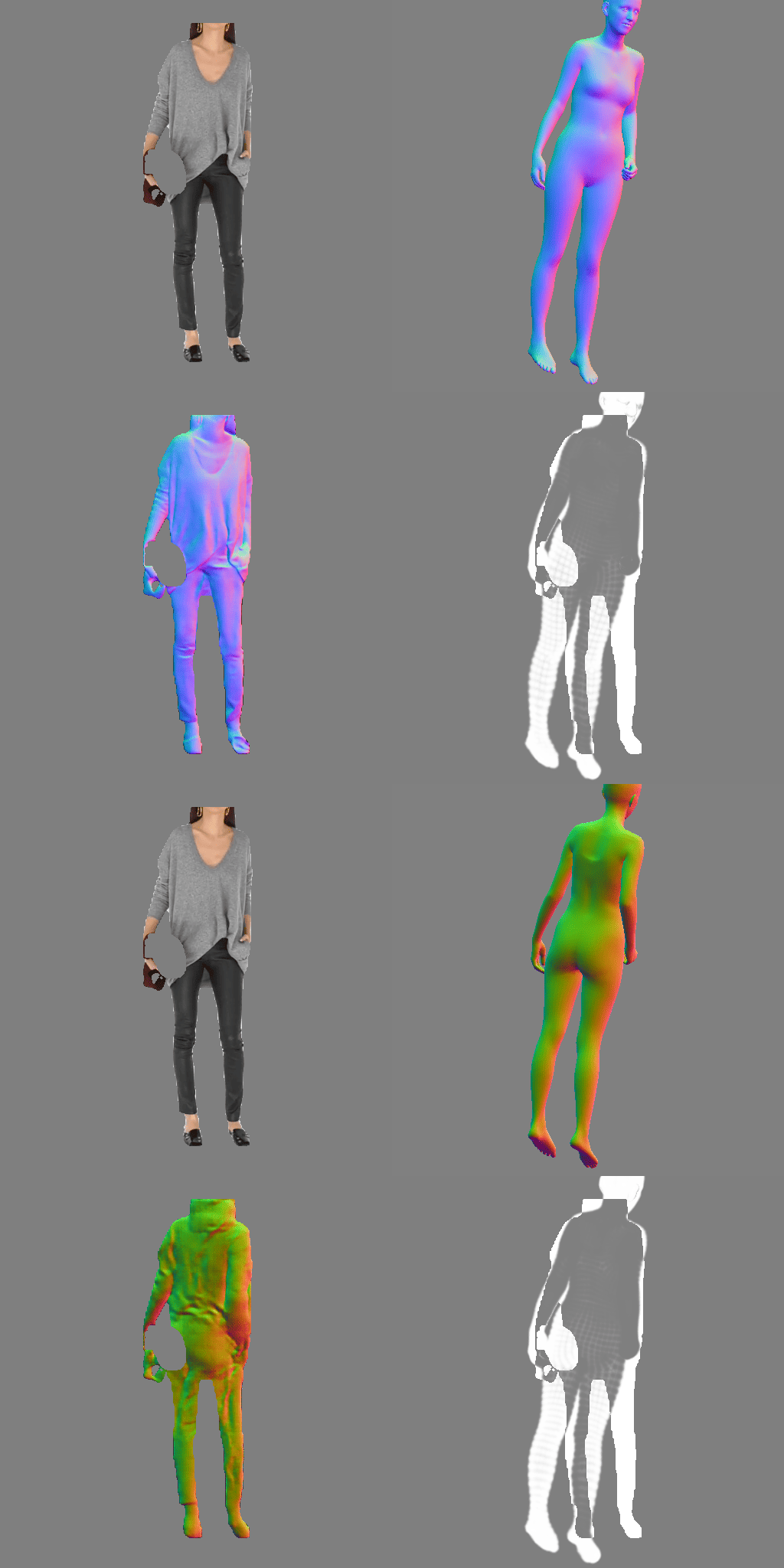}&
        \includegraphics[width=1.08in]{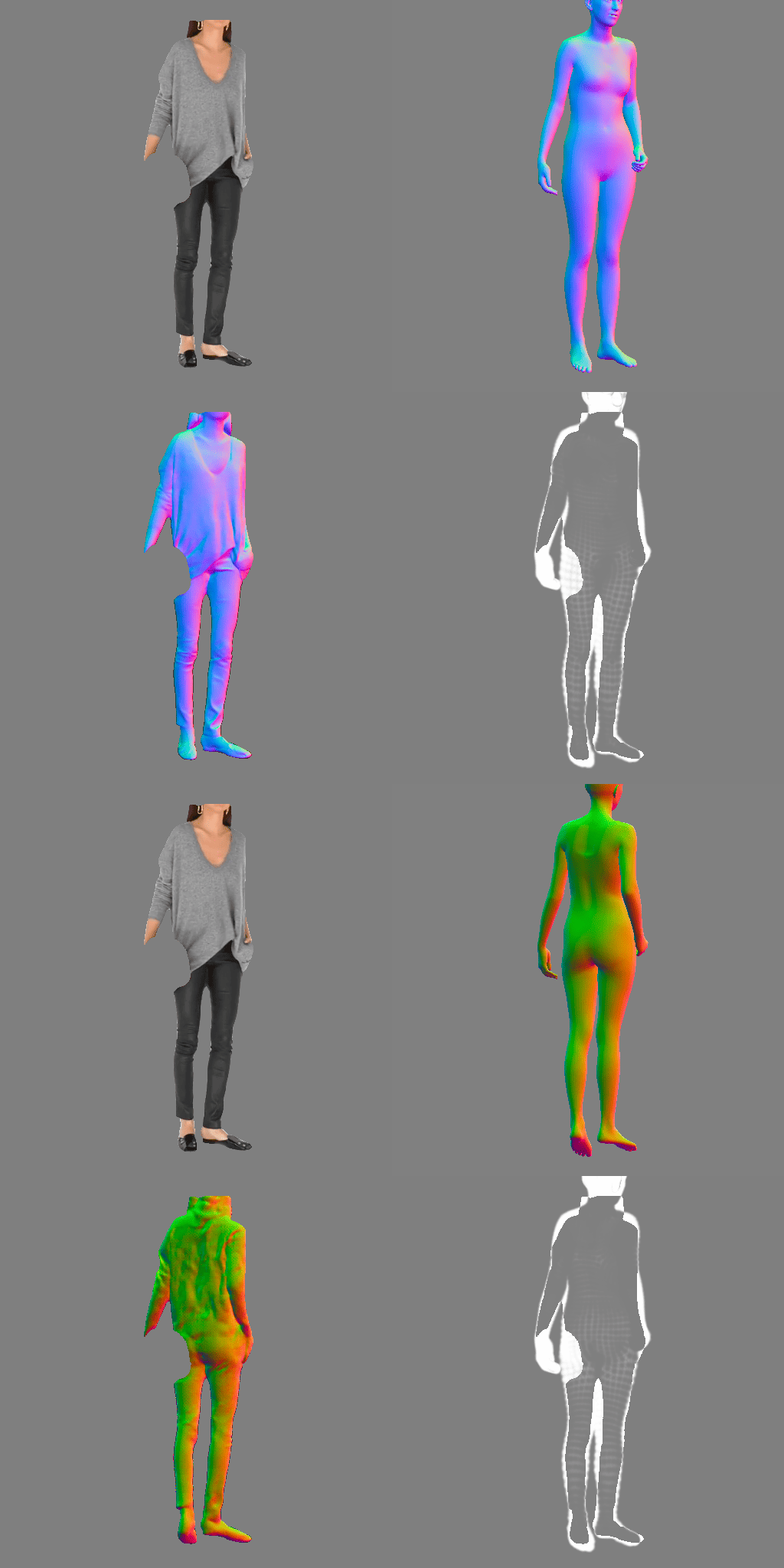}&

        \includegraphics[width=1.08in]{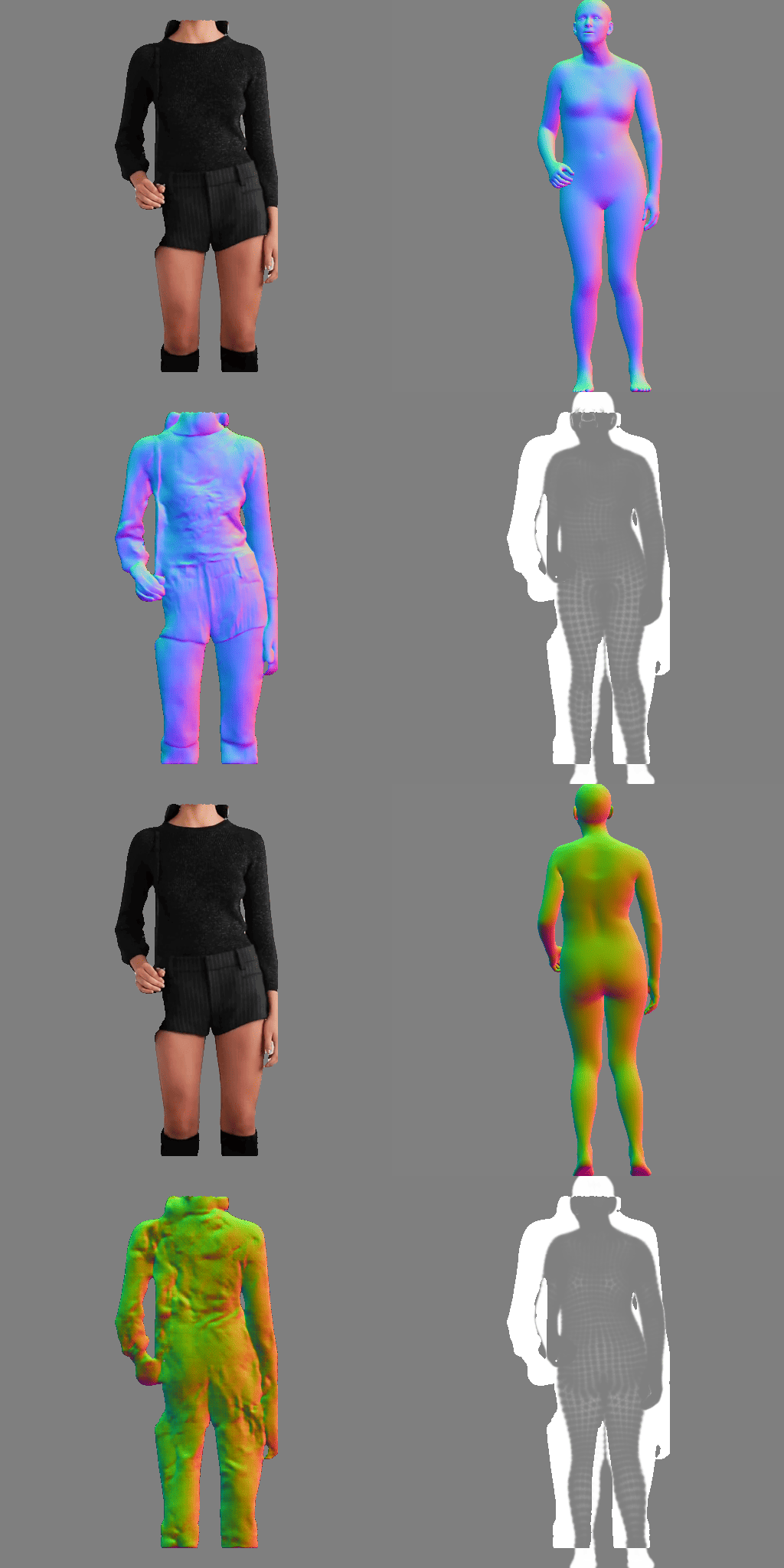}&
        \includegraphics[width=1.08in]{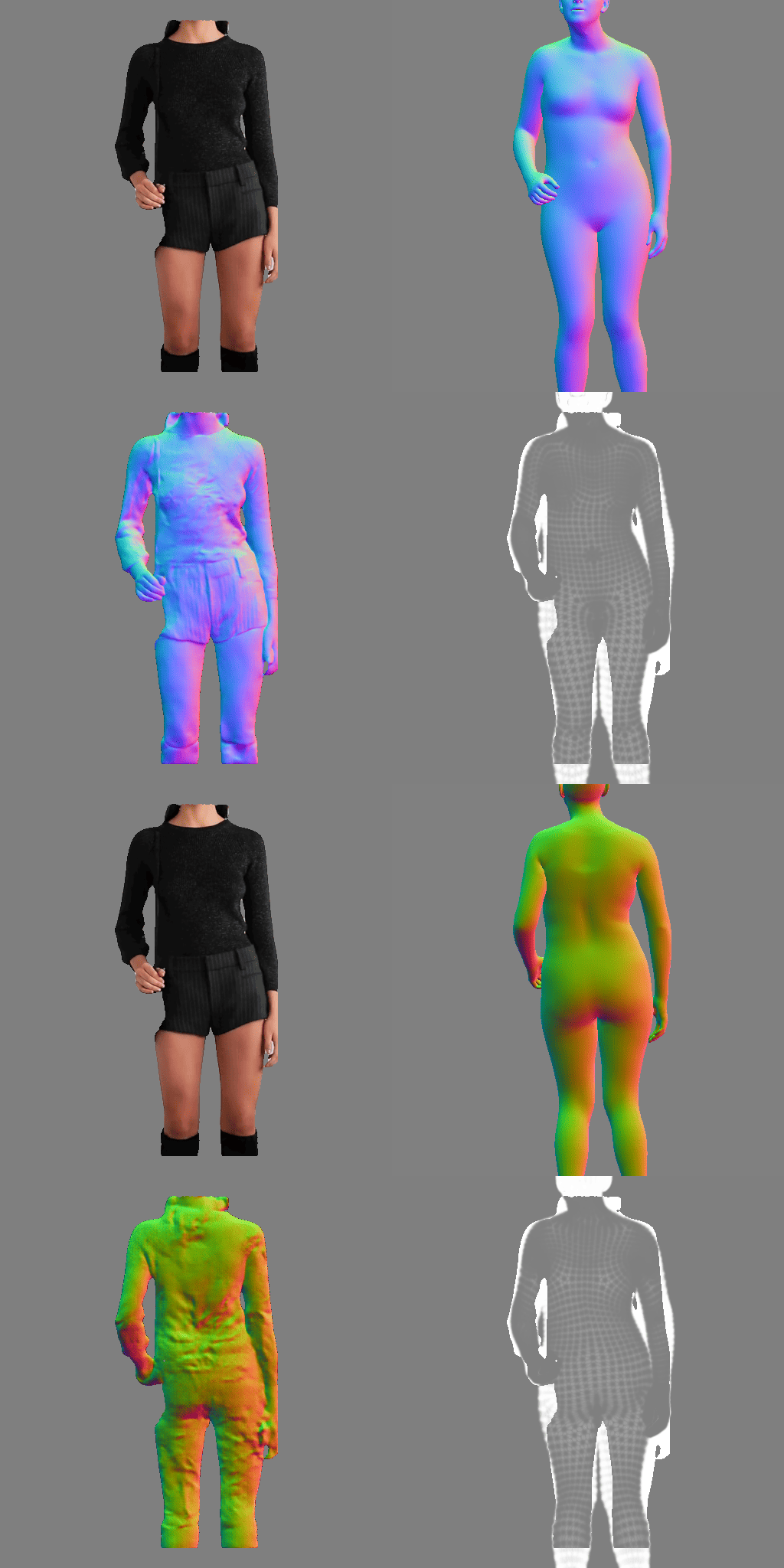}&

        \includegraphics[width=1.08in]{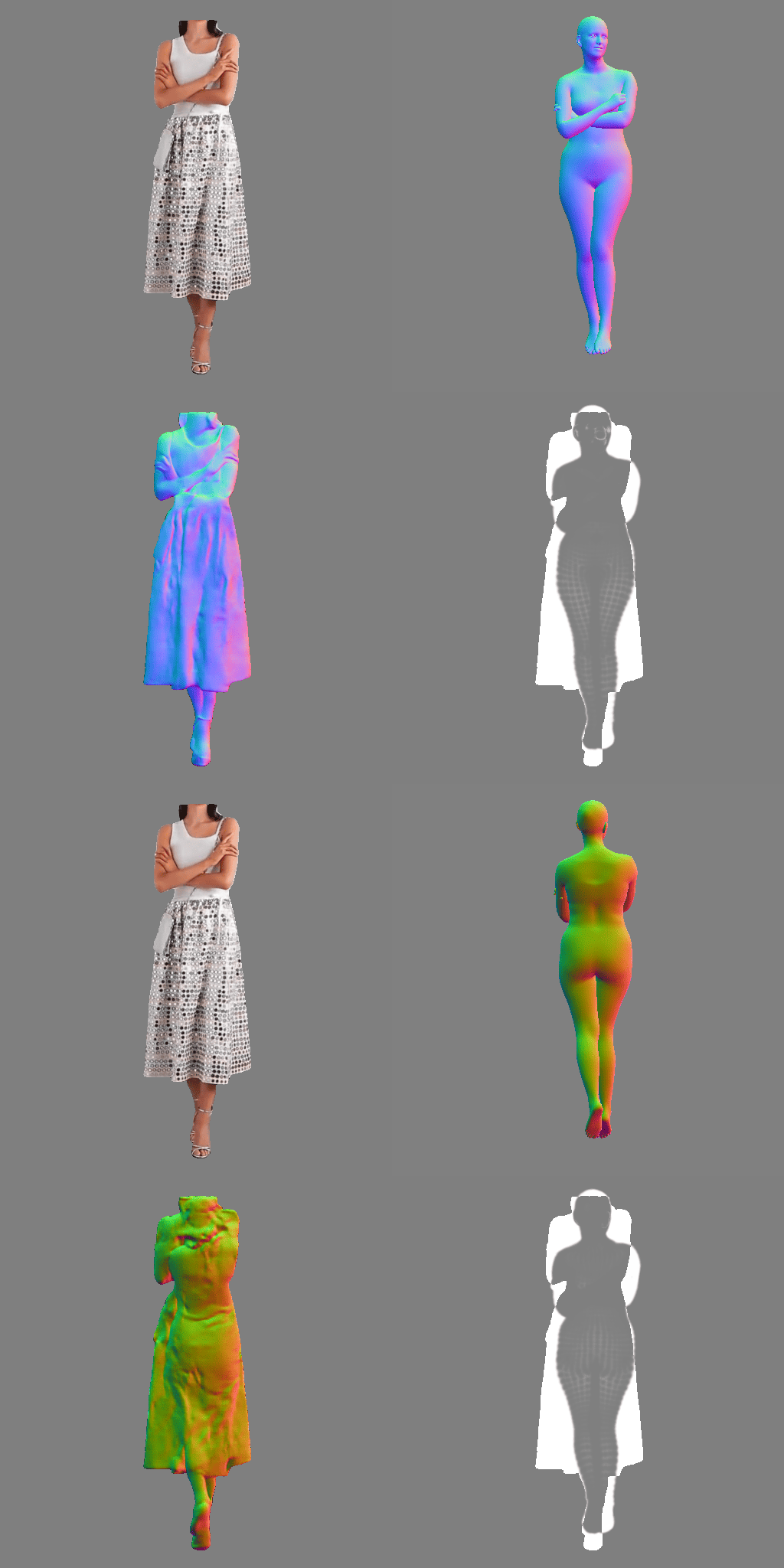}&
        \includegraphics[width=1.08in]{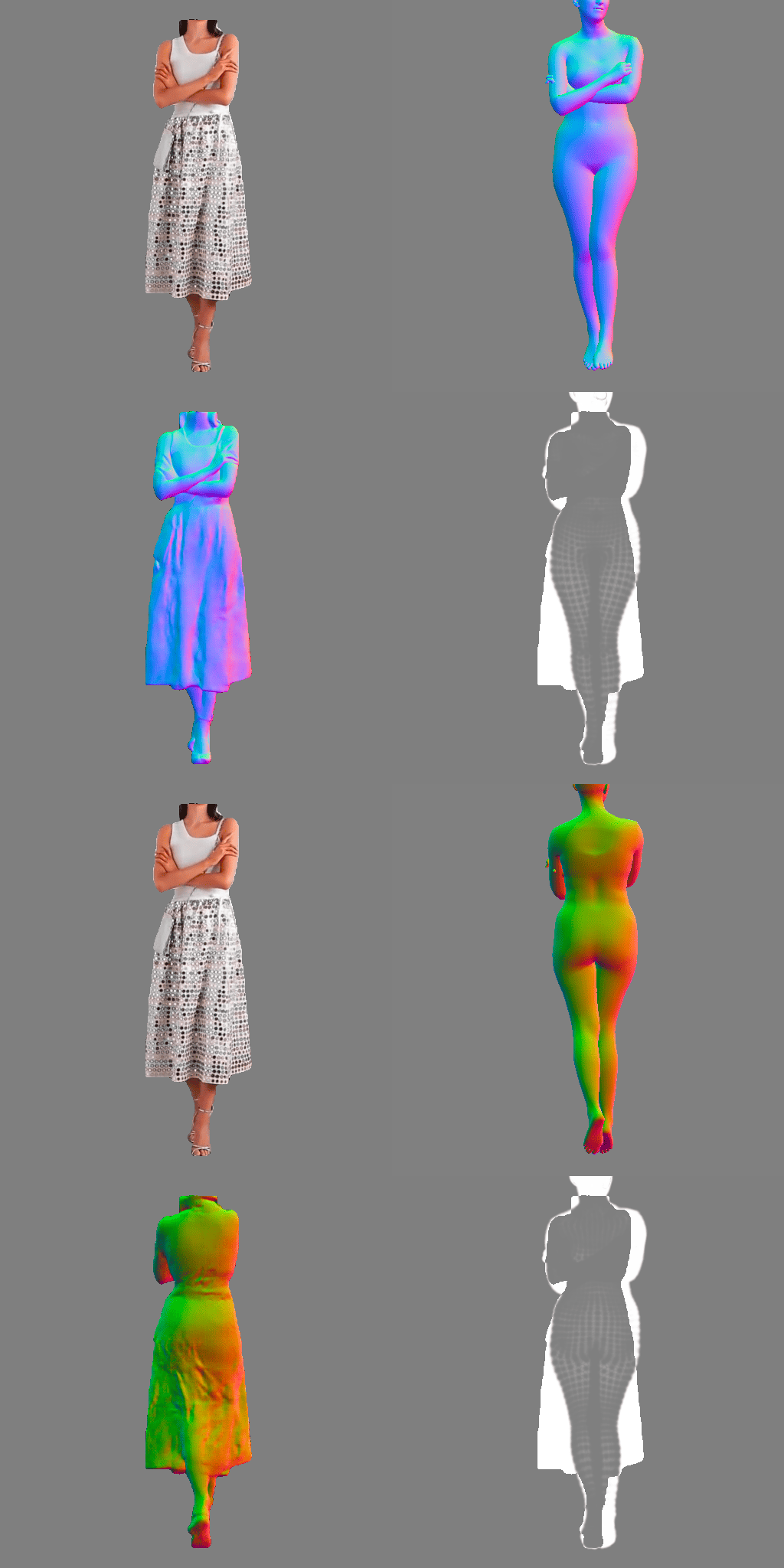}\\
        \vspace{-0.2in}\\
        
        before & after & before & after & before & after\\
        \vspace{-0.25in}\\
        \multicolumn{2}{c}{(a)} & \multicolumn{2}{c}{(b)} & \multicolumn{2}{c}{(c)} \\
        
    \end{tabular}
    \vspace{-.15in}
    \caption{\textbf{Effectiveness of our SMPL-X Refinement.}}
    \label{fig:smplx}
\end{figure*}

\begin{figure*}[h]
    \centering
    \tabcolsep=0.02in
    \begin{tabular}{c}
        \includegraphics[width=6.6in]{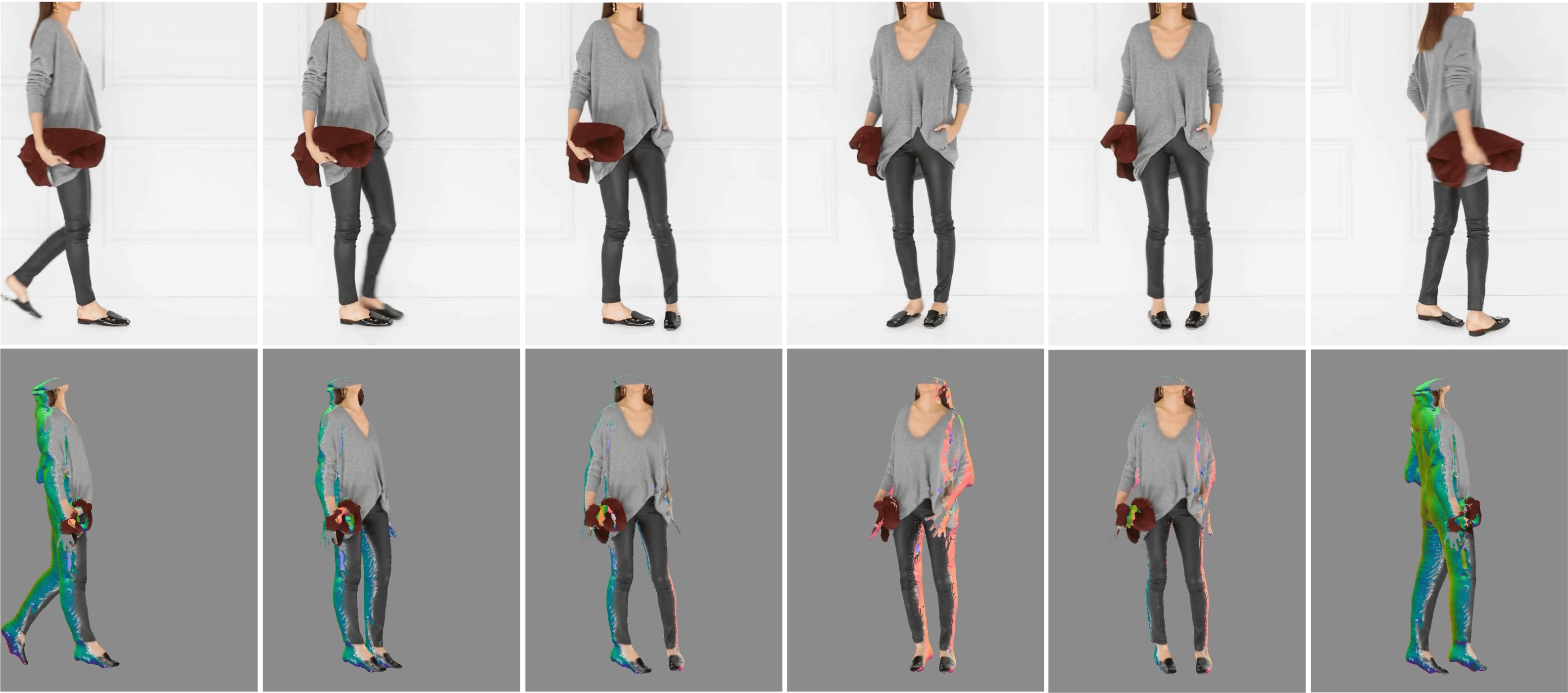}\\
        \includegraphics[width=6.6in]{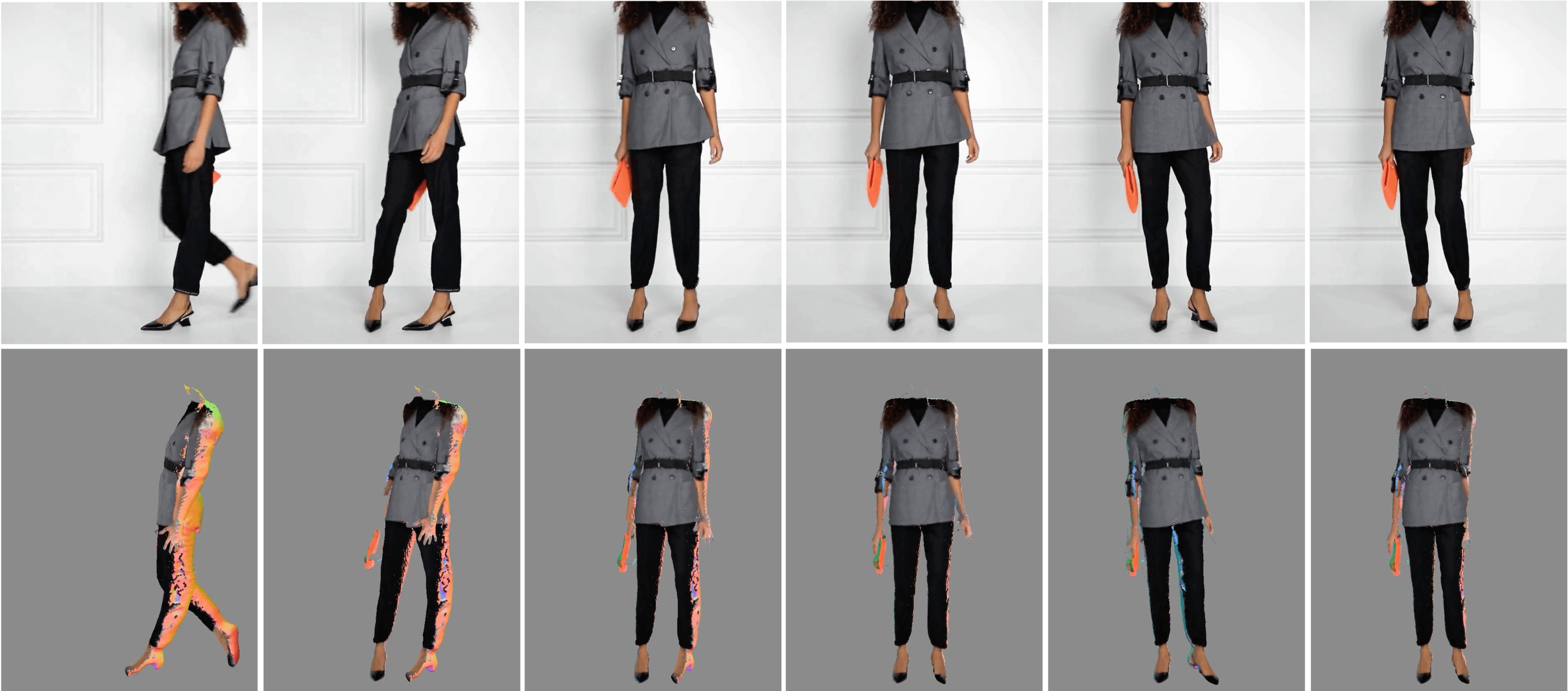}\\
        \includegraphics[width=6.6in]{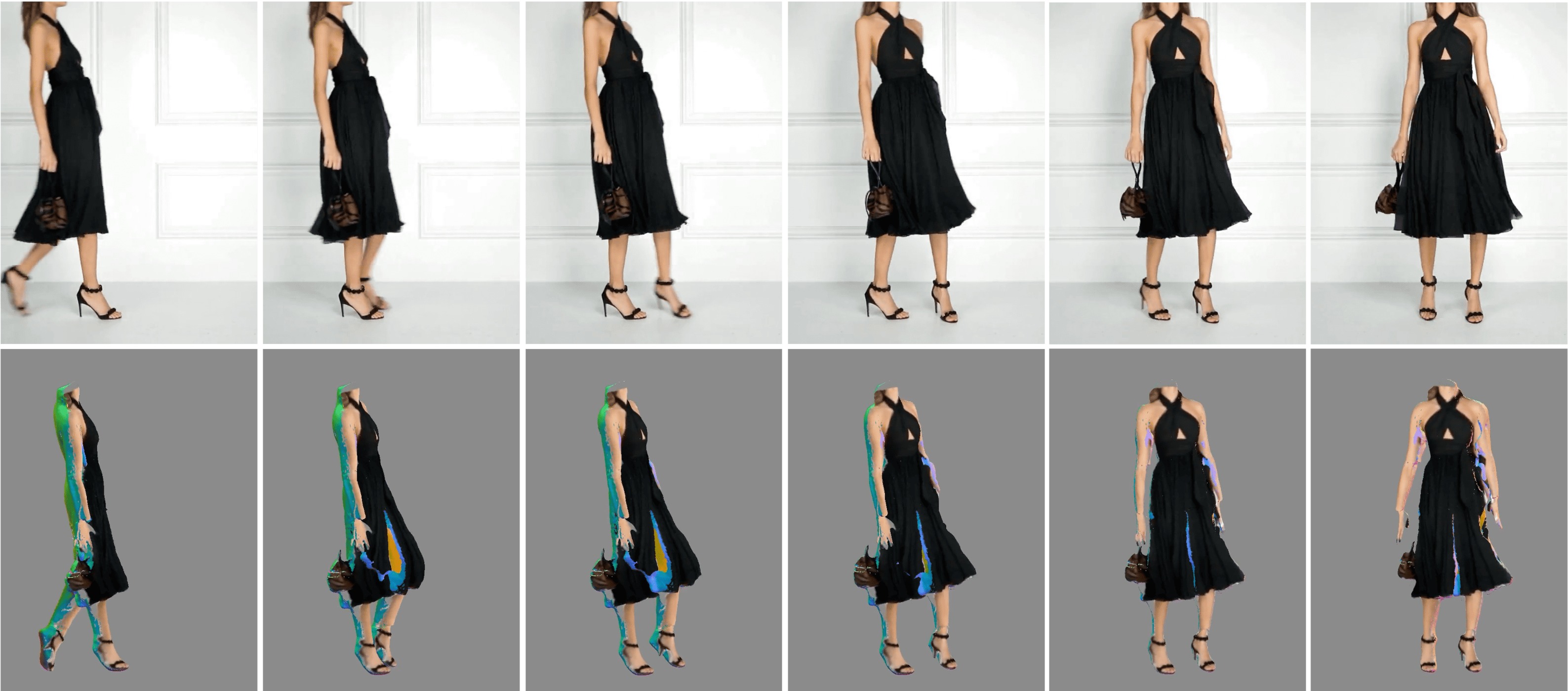}\\        
    \end{tabular}
    \vspace{-.15in}
    \caption{\textbf{Animated 3D guidance.}}
    \label{fig:3d}
\end{figure*}

As shown in Figure~\ref{fig:smplx}, if we optimize the pose of SMPL-X in Panel (a), the pose refinement may be abnormal due to the incomplete human body parts, leading to reconstruction failure. Thanks to the powerful generative capabilities of the diffusion model~\cite{ddpm,sd}, which do not require high precision for the pose accuracy in 3D guidance, we choose to freeze the pose parameters $\theta$, as this approach is sufficient to yield usable results for the robust 3D guidance reconstruction. In Panel (b) and (c), current HPS regressors often yield inaccurate camera scale estimations. To address this issue, we simultaneously optimize the camera scale $s$ applied to SMPL-X. However, for incomplete human bodies, the camera scale $s$ tends to be abnormally reduced. We use a unidirectional regularization penalty to constrain the optimization direction of the $s$. Figure~\ref{fig:3d} demonstrates that our 3D pipeline is applicable to most scenarios.

\begin{figure*}[!t]
    \centering
    \tabcolsep=0.02in
    \begin{tabular}{cc}
        \includegraphics[width=2.8in]{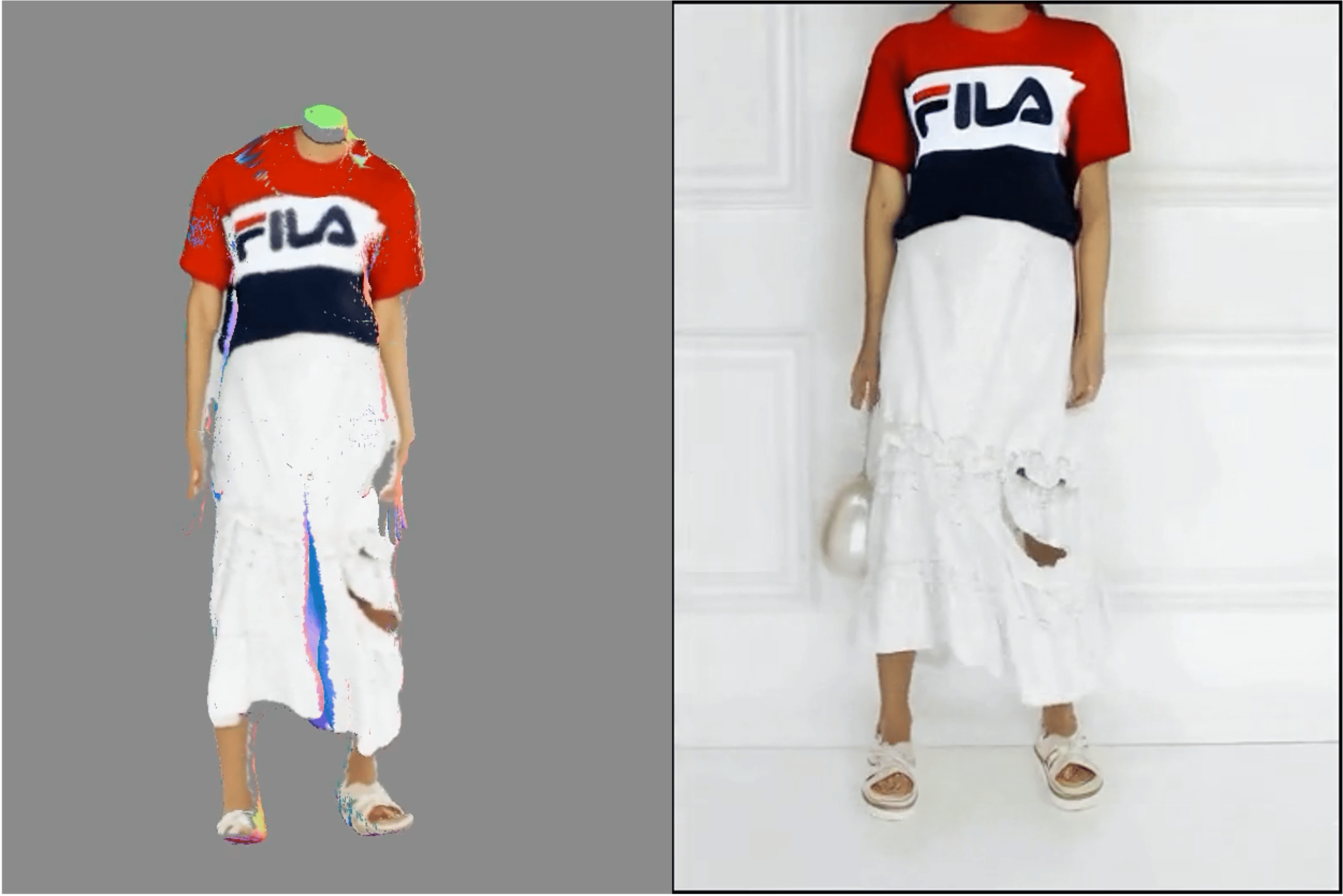}&
        \includegraphics[width=2.8in]{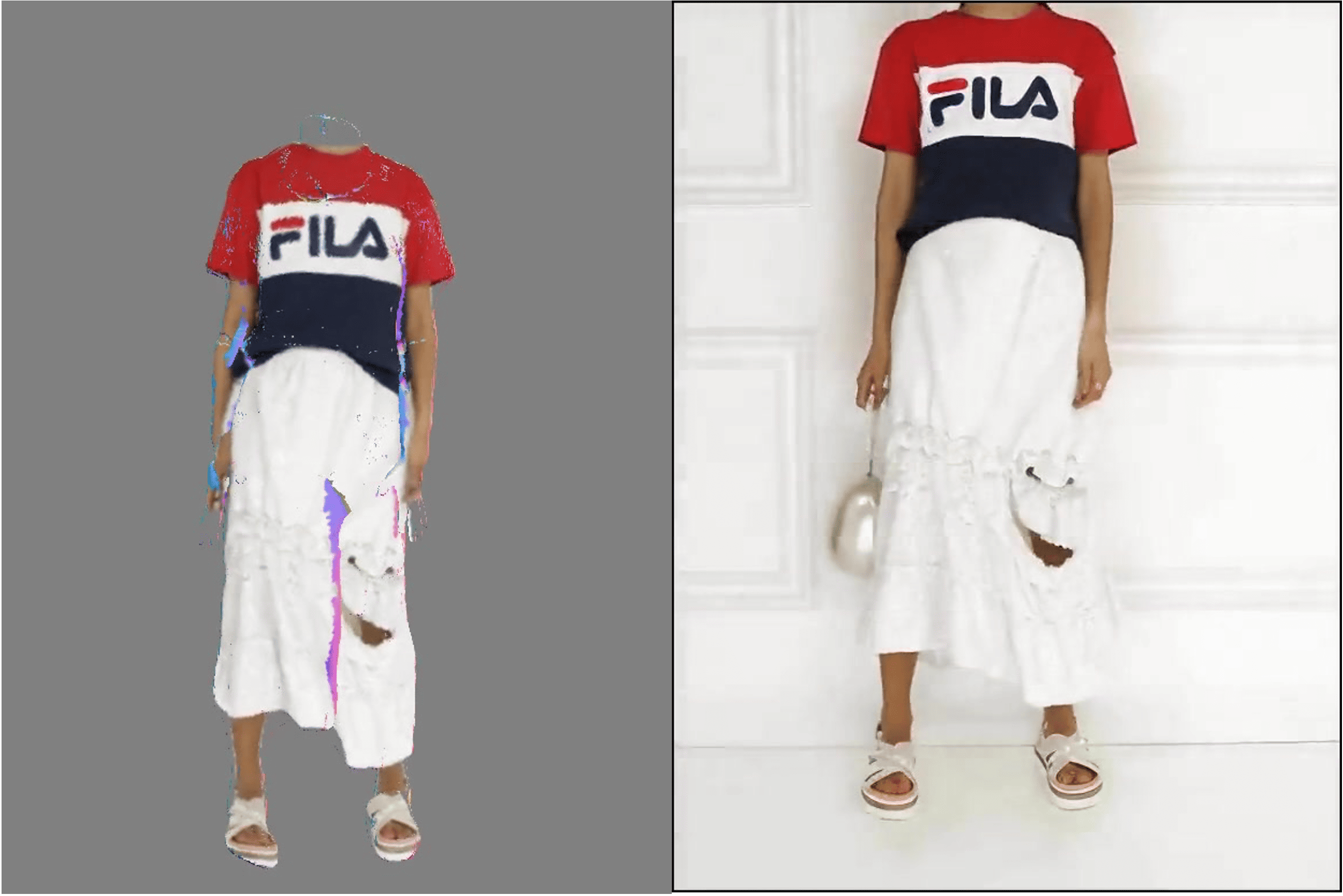}\\
    \end{tabular}
    \vspace{-.1in}
    \caption{\textbf{Effectiveness of our textured 3D animation method.}}
    \label{fig:animation}
\end{figure*}

\begin{figure*}[!t]
    \centering
    \tabcolsep=0.02in
    \begin{tabular}{c}
        \includegraphics[width=6.5in]{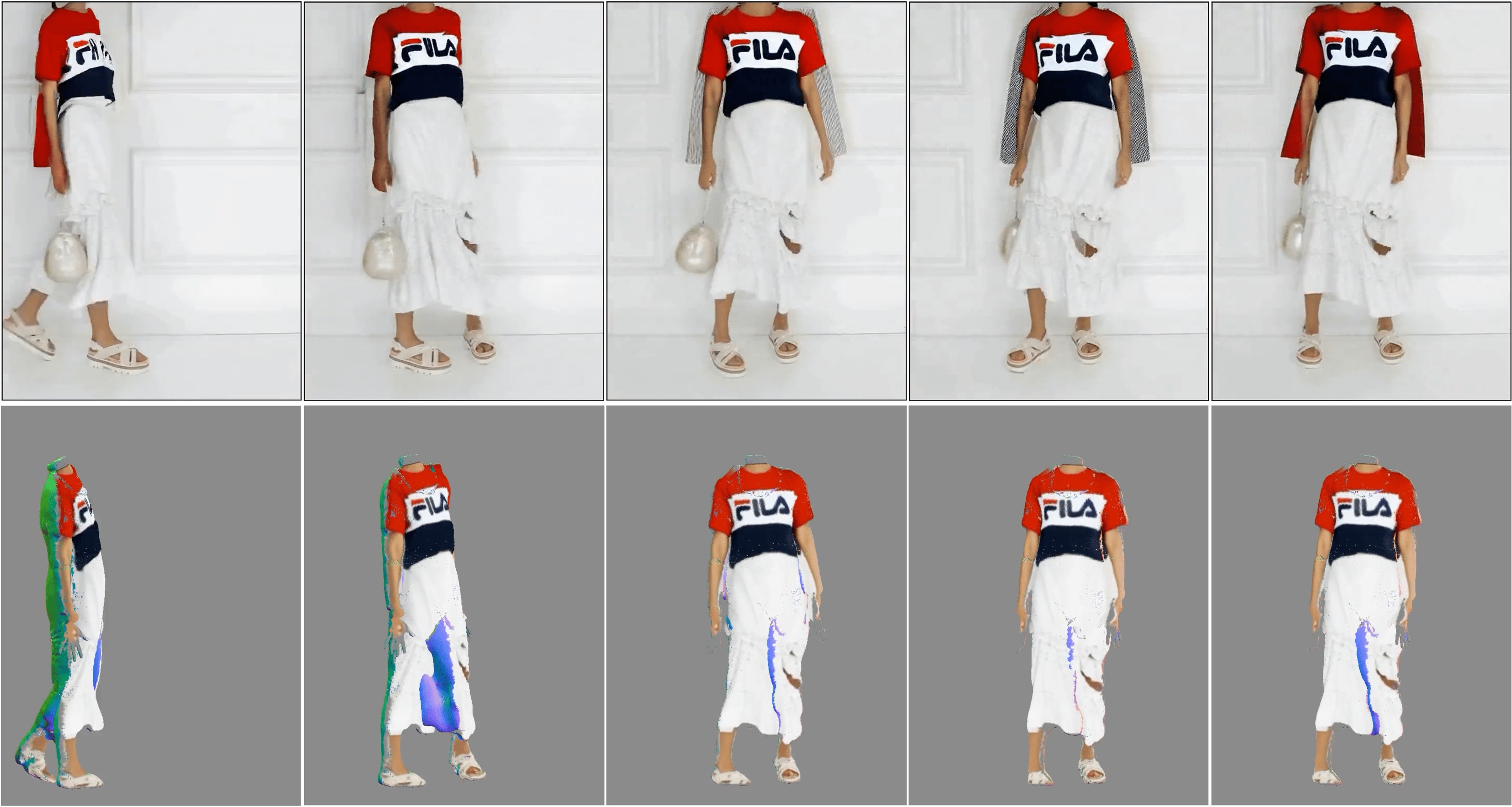}\\    
    \end{tabular}
    \vspace{-.1in}
    \caption{\textbf{Effectiveness of freezing temporal attention of our 3DV-TON.}}
    \label{fig:notem}
\end{figure*}

\noindent\textbf{Animation.}
To ensure smooth animation of the 3D guidance, we employ a video-based SMPL estimation approach~\cite{wham,gvhmr}. However, during the reconstruction phase, our input is an image, and to achieve more accurate reconstruction, we employ an image-based SMPL-X estimation method. Since there are differences in shape and other parameters between the body estimations from video and image-based methods, directly using the body sequences estimated from video for animating may result in texture distortion and deformation, as shown in Figure~\ref{fig:animation} (left). To address this issue, we utilize the video-based estimated SMPL for rigging the reconstructed clothed human before animation. Figure~\ref{fig:animation} (right) demonstrates that our method effectively avoids texture distortion and deformation.

\section{Network Architecture}
\noindent\textbf{Temporal attention.} Since our textured 3D guidance provides sufficiently explicit frame-level references, we find that our 3DV-TON can maintain texture consistency even when temporal attention is freezed (initialized with AnimateDiff~\cite{animatediff}), albeit with some minor jitter and mask aritfacts. Texture errors occur only when the 3D guidance hard to provide texture references, as shown in the first two columns of Figure~\ref{fig:notem}. This demonstrates that our 3DV-TON, using textured 3D guidance, is capable of generating consistent texture motion rather than overly focusing on smoothing inconsistent content between frames.

\section{More Results}
\begin{figure*}[!t]
    \centering
    \tabcolsep=0.02in
    \begin{tabular}{c}
        \includegraphics[width=6.8in]{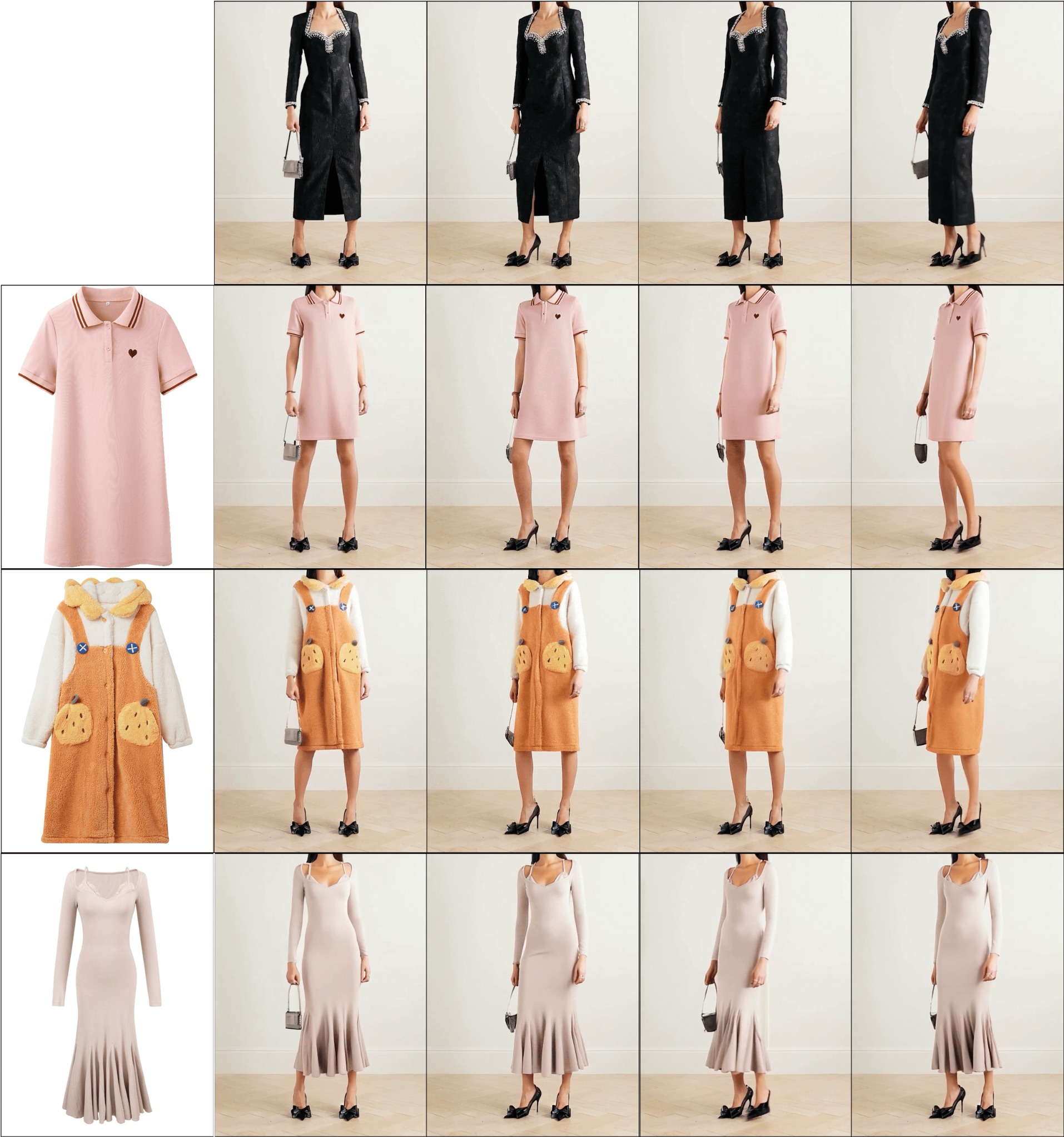}\\    
    \end{tabular}
    \vspace{-.1in}
    \caption{\textbf{More results generated by 3DV-TON.}}
    \label{fig:res1}
\end{figure*}

As shown in Figure~\ref{fig:res1}, our method can handle various shape, materials, and complex textures of clothing, while generating consistent texture motions. 

For more qualitative comparisons and video try-on results, please refer to the project page.

\end{document}